\documentclass[10pt,twocolumn,letterpaper]{article}

\usepackage[pagenumbers]{cvpr} 
\usepackage{float}

\usepackage{xcolor}

\definecolor{cvprblue}{rgb}{0.21,0.49,0.74}
\usepackage[pagebackref,breaklinks,colorlinks,allcolors=cvprblue]{hyperref}
\def\paperID{35130} 
\def\confName{CVPR}
\def\confYear{2026}

\title{BuildAnyPoint: 3D Building Structured Abstraction from Diverse Point Clouds}

\author{Tongyan Hua$^{\dag 1}$ \quad 
Haoran Gong$^{\dag 2}$\quad 
Yuan Liu$^{3}$\quad 
Di Wang$^{2}$\quad 
Ying-Cong Chen$^{1,3}$\quad 
Wufan Zhao$^{\ast 1}$ \\
$^{1}$HKUST(GZ) \quad $^{2}$Xi’an Jiaotong University \quad $^{3}$HKUST  \\
{\tt\small  thua388@connect.hkust-gz.edu.cn \quad
gonghr@stu.xjtu.edu.cn \quad diwang@mail.xjtu.edu.cn} \\
{\tt\small
\{yuanly, yingcongchen\}@ust.hk \quad
wufanzhao@hkust-gz.edu.cn}\\
\small{Project Page: \url{https://ai4city-hkust.github.io/BuildAnyPoint/}}
}

\begin{document}
\include{macro}

\twocolumn[{%
\renewcommand\twocolumn[1][]{#1}%
\maketitle

\thispagestyle{empty}
\begin{center}
    \includegraphics[width=1\textwidth]{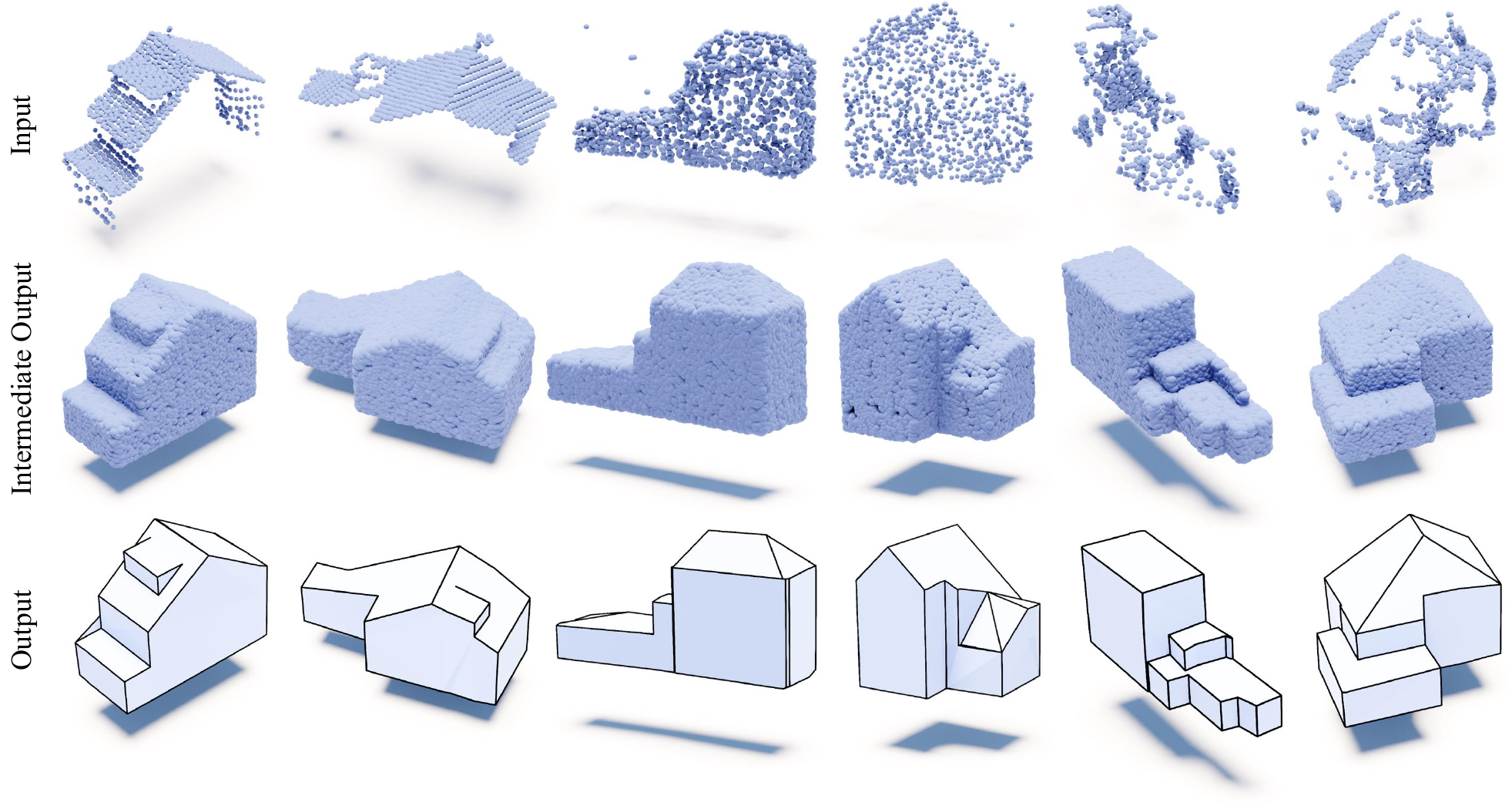}
    \vspace{-6mm}
    \captionof{figure}{\textbf{BuildAnyPoint} showcases remarkable generalization across various point cloud distributions commonly found in urban settings. Left: airborne LiDAR. Middle: Structure-from-Motion. Right: noisy sparse sampling. 
    }
    \label{fig:fig1_teaser}
\end{center}
}]
\begingroup
\renewcommand\thefootnote{}
\footnotetext{
$^{\dag}$ Equal contribution. 
$^{\ast}$ Corresponding author.}
\endgroup



\begin{abstract}

We introduce \textbf{BuildAnyPoint}, a novel generative framework for structured 3D building reconstruction from point clouds with diverse distributions, such as those captured by airborne LiDAR and Structure-from-Motion.
To recover artist-created building abstraction in this highly underconstrained setting, we capitalize on the role of explicit 3D generative priors in autoregressive mesh generation.
Specifically, we design a Loosely Cascaded Diffusion Transformer (\textbf{Loca-DiT}) that initially recovers the underlying distribution from noisy or sparse points, followed by autoregressively encapsulating them into compact meshes.
We first formulate distribution recovery as a conditional generation task by training latent diffusion models conditioned on input point clouds, and then tailor a decoder-only transformer for conditional autoregressive mesh generation based on the recovered point clouds.
Our method delivers substantial qualitative and quantitative improvements over prior building abstraction methods. Furthermore, the effectiveness of our approach is evidenced by the strong performance of its recovered point clouds on building point cloud completion benchmarks, which exhibit improved surface accuracy and distribution uniformity.




\end{abstract}

\section{Introduction}
\label{sec:intro}

Retrieving 3D building structure abstractions from unstructured observations of urban environments, primarily captured as point clouds, is crucial for a wide variety of applications, including visualization as digital twins~\cite{pan2025building,verdie2015lod}, navigation~\cite{zhu2024lod}, disaster simulation~\cite{nagasawa2021model}, planning, and general immersive experience~\cite{biljecki2015applications}.
Therefore, various efforts, including optimization-based~\cite{bauchet2020kinetic,nan2017polyfit,huang2022city3d} and learning-based~\cite{chen2022reconstructing,li2022point2roof,chen2024polygnn,liu2024point2building} approaches, have been developed, typically to handle point clouds with specific distributions, such as dense photogrammetric reconstruction~\cite{pan2025building,nan2017polyfit}, sparse observations from large-scale airborne LiDAR~\cite{bauchet2024simplicity,liu2024point2building,huang2022city3d,wang2023building3d} and Structure-from-Motion (SfM)~\cite{huang2025arcpro}.
Unfortunately, none of these methods is capable of handling arbitrary point cloud distributions, thereby significantly limiting the broader applicability of this technology~\cite{yuan2025immersegen}.

By pioneering direct, autoregressive mesh generation from point clouds, Point2Building~\cite{liu2024point2building} enhances adaptability to diverse geometries, eliminating the dependency on the plane detection or other pre-processing steps common in earlier works~\cite{nan2017polyfit,yang2022connectivity,chen2024polygnn}.
Nevertheless, its reliance on a single autoregressive step often yields geometrically ambiguous results and suboptimal mesh-point cloud alignment.
%
%
Bridging this gap, an intermediate representation between point clouds and meshes is initiated in recent work~\cite{huang2025arcpro} to reduce ambiguity by restricting the solution space to a set of valid
architectural grammar. 
This improved alignment enforces structural coherence, which regularizes the ill-posed reconstruction problem and thereby enhances generalization to unseen layouts while suppressing overfitting to input artifacts.
However, this advantage comes at the expense of structural flexibility, as the solution is confined to predefined geometry, limiting its ability to accommodate more complex structures. More critically, this design incorporates a strong structural prior, assuming that each building module is supported by relatively complete local point clouds for its generation. This assumption may pose challenges for point clouds with extreme sparsity and irregular density, such as those captured by airborne LiDAR.

In this work, we demonstrate that the underlying distribution of 3D buildings can be explicitly recovered from heterogeneous point cloud observations, eliminating the requirement for hand-crafted external modalities such as architectural grammars.
The resulting dense, uniform point clouds emulate the high-quality inputs required by state-of-the-art autoregressive mesh generators~\cite{chen2024meshanything, chen2024meshanythingv2, hao2024meshtron}, thereby enabling the retention of high-fidelity geometric detail.
%
Our key insight is to leverage an explicit 3D generative prior that constrains the solution space to a distribution of plausible shapes.
This probabilistic formulation inherently guards against overfitting to sparse or noisy inputs and generalizes to a broader class of geometries than those defined by a fixed grammar.

%
%

\begin{figure}[t!]
\centering
\includegraphics[width=1\linewidth]{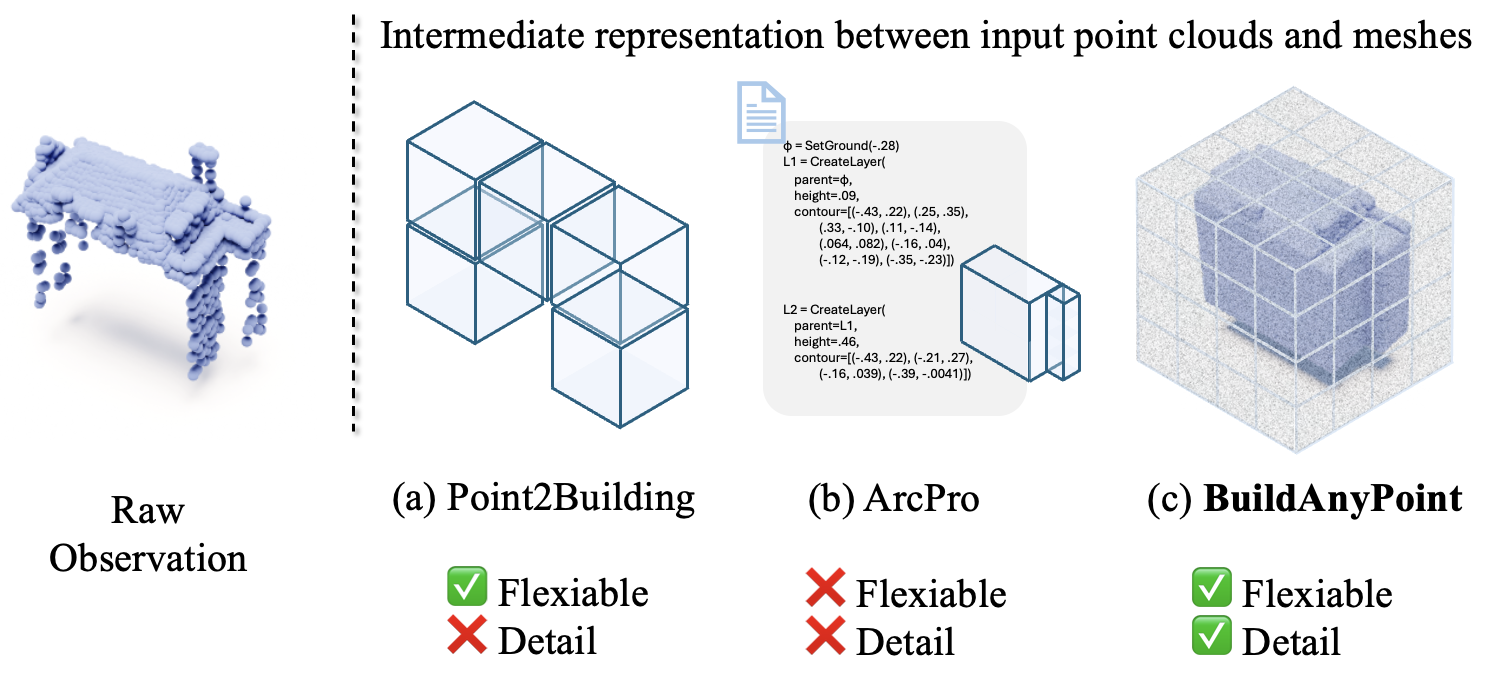}
\vspace{-3mm}
\caption{
(a) Point2Building~\cite{liu2024point2building} aggressively compresses point clouds into a low-resolution feature grid to compensate for missing regions in the raw inputs.
(b) ArcPro~\cite{huang2025arcpro} is constrained by a fixed architectural grammar that can only generate vertically extruded geometric primitives, limiting flexibility in representing common structures such as slanted roofs.
(c) \textbf{BuildAnyPoint} probabilistically models the diverse building distributions within a high-resolution latent grid.
}
\label{fig:intro}
\vspace{-6mm}
\end{figure}

Our primary objective is to develope a generalizable framework for reconstructing 3D structured buildings from diverse point cloud distributions. To achieve this ambitious goal, we design a Loosely Cascaded Diffusion Transformer, termed \emph{Loca-DiT}, as an engine to progressively model the 3D generative prior and subsequently transform it into sequential tokens that represent 3D buildings. 
Loca-DiT consists of two key stages: first, a hierarchical 3D latent diffusion process is utilized to robustly recover the building’s geometric prior from noisy point clouds, generating an intermediate representation that can be converted into dense point clouds.
Subsequently, this representation serves as a conditional input for an autoregressive Transformer, which generates high-quality, low-polygon, and topologically consistent meshes. 
This loosely cascaded paradigm bridges the modality gap in latent space, transitioning from explicit feature grids suited to unstructured point inputs to sequential tokens for structured mesh reconstruction.



Our work builds upon the recent advancements in 3D latent diffusion and autoregressive mesh generation models~\cite{ren2024xcube,chen2024meshanythingv2}. 
Extensive experiments on diverse point cloud distributions show that our method achieves superior 3D mesh quality compared to state-of-the-art approaches. Furthermore, its leading performance on building point cloud completion benchmarks demonstrates strong potential for broader application in related domains.
The main contributions of our work are summarized as follows:

\begin{itemize}
    \item The first generalizable framework that achieves 3D building structured abstraction across diverse point cloud distributions. This is achieved by directly recovering dense, accurate, and uniformly distributed point clouds from challenging sources such as airborne LiDAR and SfM, without domain-specific pre-processing.
    \item A novel loosely-cascaded architecture that synergizes hierarchical sparse diffusion with autoregressive sequence modeling to facilitate a smooth, progressive alignment of modalities through a series of latent space transitions.
    \item State-of-the-art performance in both final and intermediate product quality, further validated by leading results on dedicated building completion benchmarks.
\end{itemize}


%
%


\section{Related Work}
\label{sec:re}

\subsection{3D Shape Generation}
The field of 3D object generation is undergoing a significant shift. Initially reliant on distilling 2D knowledge from image foundation models~\cite{2dlin2023magic3d,2dwang2023prolificdreamer,2dpoole2022dreamfusion}, research is now moving toward native 3D diffusion models~\cite{3dxiang2025structured,3dzhang2024clay,3dzhang20233dshape2vecset}. This new direction prioritizes computational efficiency and the ability to create diverse, novel 3D assets from conditional inputs.
Various 3D representations, such as point clouds~\cite{pcluo2021diffusion,pcnichol2022point}, voxel grids~\cite{gridhui2022neural,gridtang2023volumediffusion}, and implicit neural representations~\cite{nigupta20233dgen,niliu2023one,niwang2023rodin} have been explored in this context.
Recently, 3D generation has increasingly adopted VAE-based latent sampling using diffusion~\cite{ren2024xcube}, transformers~\cite{3dxiang2025structured}, or diffusion transformers (DiT)~\cite{zhao2025hunyuan3d}, demonstrating remarkable generalizability.

\subsection{Autoregressive Mesh Generation}

Recent advances in artist-created mesh generation predominantly frame the task as a sequence modeling problem. This line of work autoregressively decodes meshes as sequences of geometric tokens, employing diverse representations such as learned vocabularies (\eg, MeshGPT~\cite{siddiqui2024meshgpt}, MeshAnything~\cite{chen2024meshanything}), explicit coordinates (\eg, MeshXL~\cite{chen2024meshxl}), and quantized triangle soups (\eg, Polydiff~\cite{alliegro2023polydiff}). Subsequent innovations have focused on enhancing structural coherence through coarse guidance (\eg, PivotMesh~\cite{weng2024pivotmesh}) and improving tokenization efficiency for scalability (\eg, MeshAnythingV2~\cite{chen2024meshanythingv2}, Meshtron~\cite{hao2024meshtron}). 
Most of these methods are capable of generating topologically consistent meshes from point clouds, but are restricted to high-quality, clean, and complete point clouds. 
Despite efforts to explore the use of point clouds obtained from real-world sensors, such as LiDAR, has been made, the results fall short due to the inherent limitations of 2D priors~\cite{he2025artist}.

\subsection{3D Building Abstraction from Point Clouds}

Reconstructing lightweight 3D models of buildings from physical data measurements has long presented a significant challenge, driving the development of numerous robust solutions, primarily centered on airborne LiDAR scans using plane detection and assembly techniques~\cite{buildbauchet2019city,buildbauchet2024simplicity,buildhuang2022city3d,buildverdie2015lod,buildzhou20122}.
Learning-based approaches have also been proposed, but generally face constraints, such as reliance on precise vertical surface inputs~\cite{chen2022reconstructing}, pre-processing for aggressive point cloud downsampling~\cite{liu2024point2building}, and architectural structural templates~\cite{huang2025arcpro}, while all being limited to processing point clouds with specific distributions. 
ArcPro~\cite{huang2025arcpro} pioneers the processing of noisy and sparse point clouds by mapping them to an architectural grammar. However, this approach achieves robustness at the cost of geometric diversity, constraining outputs to a limited set of primitives. It also implicitly assumes locally uniform point distributions, which limits its applicability to more irregular data.

\begin{figure}[t!]
  \centering
    \includegraphics[width=1\linewidth]{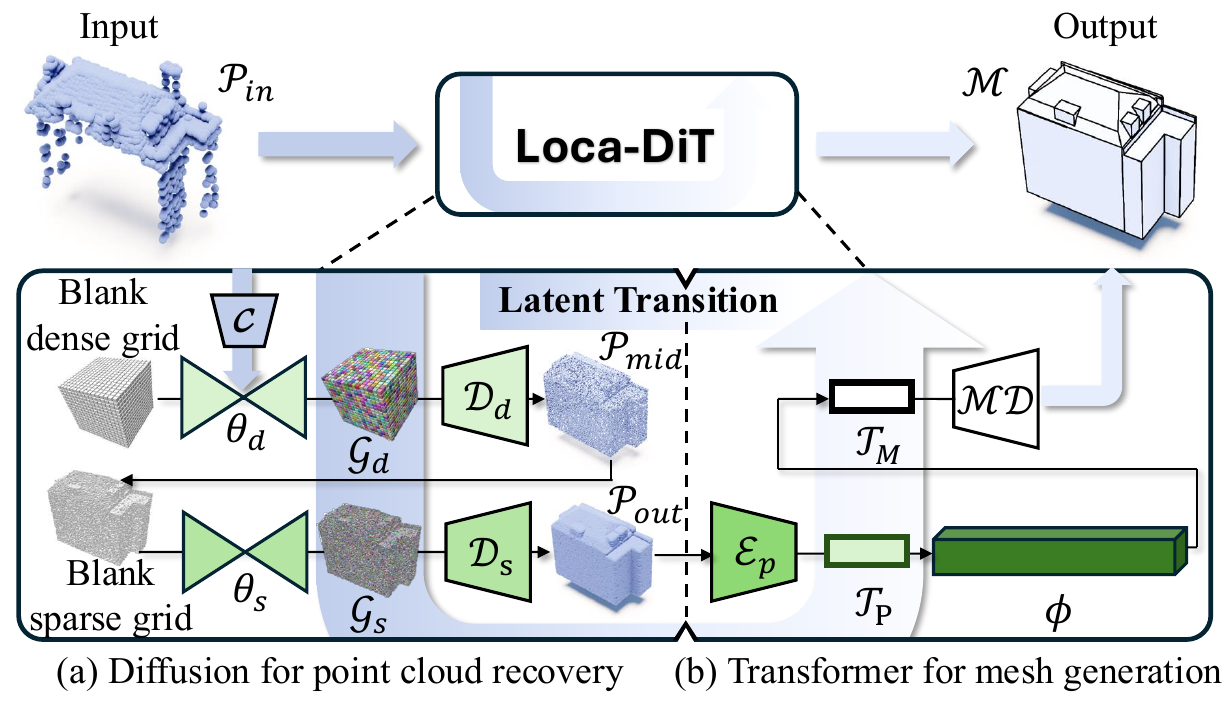}
    \vspace{-6mm}
   \caption{\textbf{Overview of BuildAnyPoint,} implemented using our generative framework Loca-DiT, which retrieves building abstraction from the input in two sequential steps via latent space transformations: (a) The hierarchical latent diffusion model $\theta$ generates an intermediate representation $\mathcal{P}_{out}$ on condition of the input point cloud $\mathcal{P}_{in}$, where the finer level of the latent representation $\mathcal{G}_s$ is conditioned on the coarser one $\mathcal{G}_d$. (b) $\mathcal{P}{out}$ is then tokenized into $\mathcal{T}{P}$ to condition a decoder-only transformer $\phi$, which autoregressively generates the mesh token sequence $\mathcal{T}_M$. The final artist-created mesh $\mathcal{M}$ is obtained by applying the Mesh Detokenization step $\mathcal{MD}$ to $\mathcal{T}_M$.}
   \label{fig:main-pipe}
   \vspace{-4mm}
\end{figure}

\section{BuildAnyPoint}

Our goal is to develop a generative framework that learns the conditional distribution \( p_{\text{BAP}}(\mathcal{M}| \mathcal{P}_{in}) \) for generating structured 3D building abstractions \(\mathcal{M}\) from input point clouds \(\mathcal{P}_{in}\) of arbitrary distributions. As illustrated in \cref{fig:main-pipe}, this is realized through a loosely cascaded generative process. The pipeline consists of a sequence of tightly coupled latent space transformations: it first encodes the input into a dense latent grid \(\mathcal{G}_{d}\), which is then refined into a sparse one \(\mathcal{G}_{s}\). 
Finally, the explicit 3D representation is serialized into a token sequence \(\mathcal{T}_P\) that serves as a prompt to guide the generation of the target mesh representation \(\mathcal{T}_M\).

\subsection{Latent Transition}
\textbf{Motivation.}
The design of distinct latent spaces is motivated by the need to bridge the significant domain gap between unstructured, potentially sparse point clouds and structured, lightweight building meshes. Closing this representational divide poses a fundamental challenge: point clouds are best encoded in continuous, dense latent spaces to capture geometric details~\cite{Feng_2022_CVPR, du2024arbitrary}, whereas meshes require discrete, sequential representations to generate structured topologies~\cite{nash2020polygen, siddiqui2024meshgpt}. Arbitrary direct alignment, however, often leads to undesirable results~\cite{he2025artist}.
We address this challenge through three dedicated stages of representation learning. First, the dense latent grid $\mathcal{G}_{d}$ recovers a complete geometric prior, handling noise and missing data. Second, the sparse latent grid $\mathcal{G}_{s}$ refines this prior into a compact and structurally optimized representation. Third, the sequential token $\mathcal{T}$ reformats the explicit 3D prior for generative sequence modeling. This deliberate separation allows each stage to specialize, enabling robust performance across diverse and challenging point cloud distributions.

\begin{figure}[t!]
\centering
\includegraphics[width=1\linewidth]{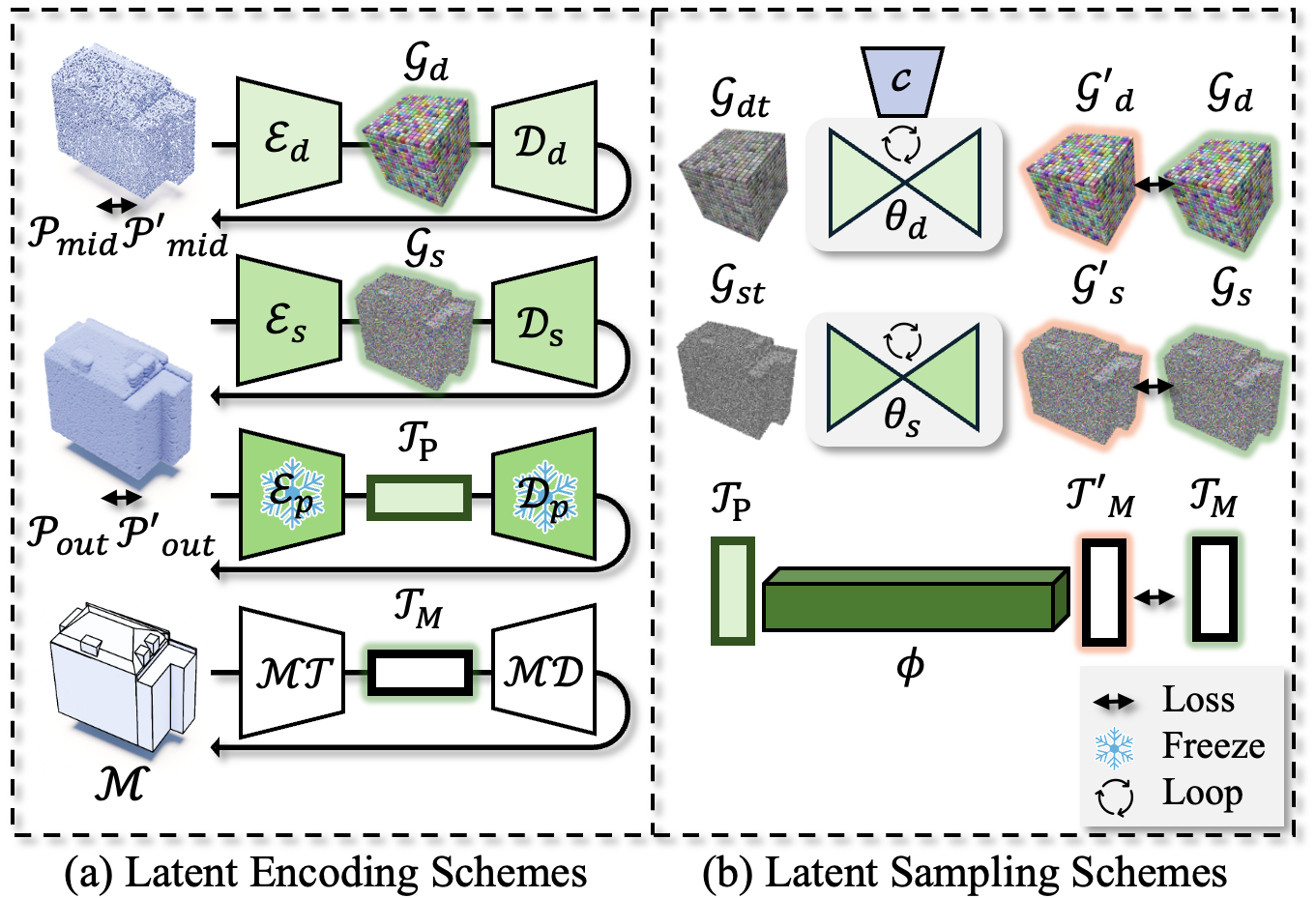}
\vspace{-6mm}
\caption{\textbf{Overview of the training process for each Loca-DiT module.} Our generative framework can be summarized as maintaining a set of latent spaces, with each tailored generative model learning how to sample and form a consistent feature space.}
\label{fig:sub-pipe}
\vspace{-6mm}
\end{figure}

\noindent\textbf{Latent encoding schemes} are designed to map each modality into a tailored latent space, thereby enhancing cross-modal alignment. This is achieved through a series of encoding and tokenization steps specified in~\cref{fig:sub-pipe}(a). 
We model the underlying point cloud distribution using points uniformly sampled from ground-truth meshes. These points are voxelized into sparse grids at low ($\mathcal{P}_{mid}$) and high ($\mathcal{P}_{out}$) resolutions, which are subsequently compressed using a separate sparse VAE structure~\cite{ren2024xcube}. 
Notably, a densification operation is applied at the bottleneck when compressing the lower-resolution sparse grid into the latent space, thereby transforming the sparse structure into a dense one, denoted as $\mathcal{G}_{d}$. This critical step provides the decoder $\mathcal{D}_d$ with complete spatial context, enabling it to reason about both occupied and unoccupied regions. As a result, during the decoding phase, which involves a series of upsampling and pruning, the model can intelligently reconstruct complete shapes by carving out unoccupied space from the dense grid.
%
%

Following the recovery of the base geometry, the refined high-resolution geometry is embedded into two latent spaces. These spaces are designed to act as mutual constraints, enforcing the alignment between the target modality and the input.
Specifically, we encode a higher-resolution voxelization of the ground-truth point cloud into an explicit 3D sparse latent grid $\mathcal{G}_{s}$ and a fixed-length ($M$) token sequence $\mathcal{T}_P$, using a sparse VAE and a pre-trained point cloud encoder~\cite{zhao2023michelangelo}, respectively. This dual encoding of the same input flattens the embodiment of explicit 3D generative priors:
{\setlength\abovedisplayskip{2pt}
\setlength\belowdisplayskip{2pt}
\begin{equation}
\mathcal{T}_P = \{t_p^1, t_p^2, \ldots, t_p^M\} = \mathcal{E}_p(\mathcal{P}_{out}),
\label{eq:token}
\end{equation}}where
$\mathcal{T}_P$ captures the nuances of the recovered underlying geometry, establishing an alignment with the target mesh token sequence $\mathcal{T}_M$. Here, $\mathcal{T}_M$ is derived from the ground-truth mesh via Mesh Tokenization ($\mathcal{MT}$), forming its compact and discretized representation~\cite{chen2024meshanythingv2}. The complete token sequence for the autoregressive model is therefore $\mathcal{T} = \{ \mathcal{T}_P, \mathcal{T}_M \}$, where the $\mathcal{T}_M$ are generated conditioned on the $\mathcal{T}_P$.

%

\subsection{Loosely Cascaded Diffusion Transformer}

The Loca-DiT, as illustrated in~\cref{fig:main-pipe}, learns the target conditional distribution $p_{\text{BAP}}(\mathcal{M}|\mathcal{P}_{in})$ by decomposing it into two probabilistically structured, yet loosely-coupled stages: 
1) A \emph{geometric completion} stage, implemented by a latent diffusion model, that recovers a complete shape $\mathcal{P}_{out} $ from the input point cloud, $p(\mathcal{P}_{out} | \mathcal{P}_{in})$;
2) A \emph{structured mesh generation} stage, implemented by an autoregressive transformer, that produces a mesh conditioned on the completed geometry, $p(\mathcal{M} | \mathcal{P}_{out})$.
Formally, the geometric completion stage prescribes the following factorization:
{\setlength\abovedisplayskip{2pt}
\setlength\belowdisplayskip{2pt}
\begin{multline}
p(\mathcal{P}_{out}|\mathcal{P}_{in}) = \iint p_{\mathcal{D}_s}(\mathcal{P}_{out}|\mathcal{G}_s) \cdot p_{\theta_s}(\mathcal{G}_s|\mathcal{G}_d) \\
\cdot p_{\theta_d}(\mathcal{G}_d|\mathcal{P}_{in}) \, d\mathcal{G}_d d\mathcal{G}_s.
\end{multline}}Here, 
\( \theta_d \) and \( \theta_s \) parameterize the hierarchical latent diffusion models responsible for sampling the dense latent grid \( \mathcal{G}_d \) and sparse latent grid \( \mathcal{G}_s \), respectively.
Meanwhile, \( \mathcal{D}_s \) represents the parameters of the sparse VAE decoder that maps the sparse latent variable \( \mathcal{G}_s \) to the target output \( \mathcal{P}_{out} \), serving as the condition for the autoregressive mesh generation stage. This process is formally factorized as:
{\setlength\abovedisplayskip{2pt}
\setlength\belowdisplayskip{2pt}
\begin{equation}
p(\mathcal{M}|\mathcal{P}_{out}) = \prod_{i=1}^{N} p_\phi(t_i|t_{<i}, \mathcal{E}_p(\mathcal{P}_{out})),
\end{equation}}where
\( \phi \) represents the parameters of the Transformer, which is responsible for autoregressively generating meshes as $N$ tokens.

\noindent\textbf{Latent sampling schemes,} such as diffusion and autoregressive transformer models, constitute a dominant paradigm in generative modeling. Despite architectural differences, they share a core principle: both learn to generate structured representations by iteratively refining a sample within a learned latent space. Starting from a prior distribution, they perform a sequence of tractable transformations, whether denoising or next-token prediction, to steer the sample towards the complex target distribution in a latent space pre-structured by our latent encoding schemes.

A diffusion process progressively transforms the target distribution into a standard Gaussian distribution by incrementally introducing Gaussian noise. In our framework, this process operates in the latent space \(\mathcal{G}\), where the forward diffusion process at timestep \(t\) perturbs the original latent variable \(\mathbf{z}_0\) as follows:
{\setlength\abovedisplayskip{2pt}
\setlength\belowdisplayskip{2pt}
\begin{equation}
\mathbf{z}_t = \sqrt{\bar{\alpha}_t} \mathbf{z}_0 + \sqrt{1-\bar{\alpha}_t} \mathbf{\epsilon},
\end{equation}}where
\(\mathbf{\epsilon} \sim \mathcal{N}(\mathbf{0}, \mathbf{I})\) and \(\bar{\alpha}_t\) is a coefficient defined by a noise schedule. The training objective is to optimize a denoising model \(\mathbf{\epsilon}_{\theta}\) that takes the noisy latent \(\mathbf{z}_t\) and the timestep \(t\) as inputs and predicts the injected noise \(\mathbf{\epsilon}\). This is achieved by minimizing the mean squared error between the predicted and true noise:
{\setlength\abovedisplayskip{2pt}
\setlength\belowdisplayskip{2pt}
\begin{equation}
\min_{\theta} \mathbb{E}_{\mathbf{z}_0 \sim \mathcal{G}, \mathbf{\epsilon} \sim \mathcal{N}(\mathbf{0}, \mathbf{I}), t \sim \mathcal{U}(1,T)} \left[ \lVert \mathbf{\epsilon} - \mathbf{\epsilon}_{\theta}( \mathbf{z}_t, t ) \rVert^2_2 \right],
\end{equation}}where
\(\mathbf{z}_0\sim \mathcal{G}\) is sampled from the latent distribution of the training set. Through this objective, \(\mathbf{\epsilon}_{\theta}\) learns to iteratively remove noise from \(\mathbf{z}_t\), recovering the clean latent \(\mathbf{z}_0\). At inference time, the trained denoising model is applied iteratively to sample a new latent variable $\mathbf{z}'_0 \sim \mathcal{G'}$ from pure Gaussian noise defined on the explicit grid structure, as shown in~\cref{fig:sub-pipe}(b).

For the transformer stage, the complete input sequence \(\mathcal{T}\) is constructed by concatenating two components, \ie, $\mathcal{T} = [\mathcal{T}_P\,;\, \mathcal{T}_M^{<t}]$, where \(\mathcal{T}_M^{<t}\) represents the partial target mesh token sequence up to timestep \(t-1\). The training objective at each step \(t\) is for the transformer model \(\phi\), conditioned on the prefix sequence \(\mathcal{T}\), to predict the next token \(t_m^{N}\) in the mesh token sequence. This corresponds to maximizing the conditional log-likelihood:
{\setlength\abovedisplayskip{2pt}
\setlength\belowdisplayskip{2pt}
\begin{equation}
\small
\max_{\phi} \log P(\mathcal{T}_M | \mathcal{T}_P) = \max_{\phi} \sum_{t=1}^{N} \log P(t_m^{N} | \mathcal{T}_P, \mathcal{T}_M^{<t}; \phi).
\label{eq:auto-train}
\end{equation}}At inference time, the trained model is used to autoregressively sample the token sequence $\mathcal{T}'_M$, given the initial point cloud tokens $\mathcal{T}_P$.


\subsection{Implementation Details}

\noindent\textbf{Latent diffusion training.} We build upon the hierarchical sparse VAE and diffusion framework outlined in XCube~\cite{ren2024xcube}. Specifically, we train the VAE and the diffusion model at the coarse and fine levels independently, as depicted in~\cref{fig:sub-pipe}(b). In the VAE training, we optimize a loss function that consists of two main components: the binary cross-entropy (BCE) between the generated and target outputs, and the Kullback-Leibler (KL) divergence between the learned posterior distribution and a prior Gaussian distribution. Our training process also incorporates the learning of vertex normals to support the autoregressive inference of mesh geometry.
The diffusion models are trained by minimizing the loss between the predicted values at each time step and a reference value derived from the VAE posterior and the noise schedule. Notably, we introduce a point encoder, acting as the condition network $c$~\cite{hua2025sat2city}, to quantize the input point cloud $\mathcal{P}_{in}$ into a voxel grid, which is then concatenated with the latent feature. Throughout the conditional diffusion process, this setup enables learning to match the predicted $\mathcal{G}'$ to the true distribution $\mathcal{G}$ provided by the VAE.


\noindent\textbf{Autoregressive transformer training.} We build our approach upon MeshAnything V2~\cite{chen2024meshanythingv2}. During training, our model leverages ground-truth point clouds and paired artist-created meshes. The meshes are converted into a compact token sequence \(\mathcal{T}_M\) via Adjacent Mesh Tokenization, while the point clouds are encoded into a conditioning token prefix \(\mathcal{T}_P\). A transformer model \(\phi\) is then trained to autoregressively predict the mesh tokens by maximizing the conditional likelihood in \cref{eq:auto-train}. At inference, the model operates conditions on a point cloud generated by a latent diffusion model, augmented with its inferred vertex normals, to synthesize a corresponding mesh that aligns with the generated geometry. Please refer to \textit{Supplementary Material} for more training details.

\begin{figure*}[th!]
  \centering
   \includegraphics[width=1\linewidth]{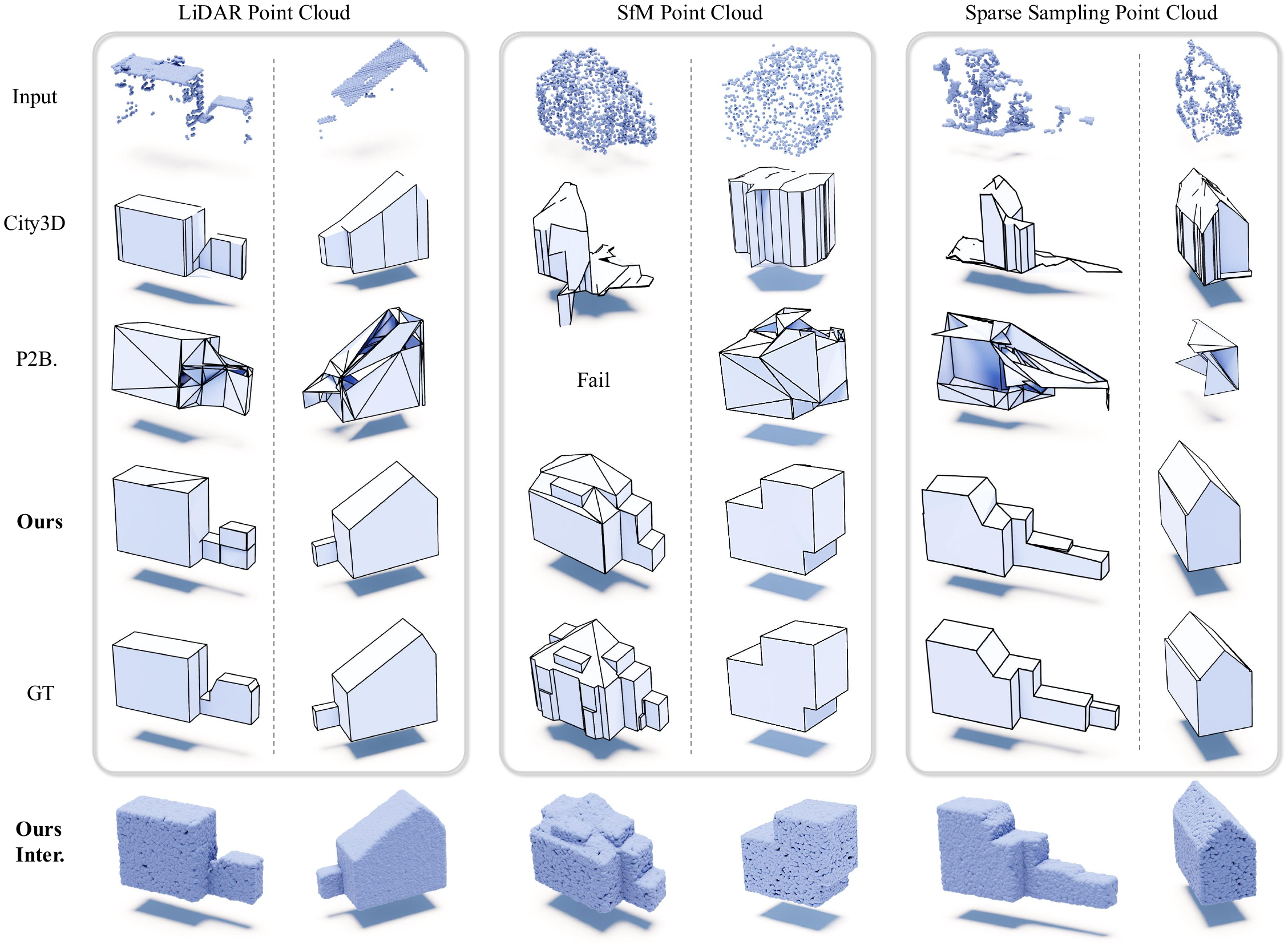}
   \vspace{-8mm}
   \caption{\textbf{3D Building Structured Abstraction Comparison.}
   Qualitative comparison on three common urban point cloud distributions against City3D~\cite{huang2022city3d} and Point2Building (abbreviated as P2B.)~\cite{liu2024point2building}. Our generative framework achieves more complete and faithful structural recovery than the alternatives, a result attributed to its robust intermediate dense points (abbreviated as Inter.) reconstructed from the 3D generative prior, which ensures consistency across heterogeneous input scenarios.}
   \label{fig:main-result}
   \vspace{-4mm}
\end{figure*}

\section{Experiments}

\subsection{Configuration}

\noindent\textbf{Datasets.}
We use the real-world benchmark which consists of 50,000 building instances from The Hague and Rotterdam, Netherlands~\cite{isprs-annals-X-4-W5-2024-179-2024}. Each instance is paired with incomplete point clouds captured by airborne LiDAR (AHN3 and AHN4 versions) and complete point clouds sampled from manually reconstructed LoD2 building models, serving as ground truth. This dataset captures the inherent incompleteness, noise, and variability of real-world point clouds, making it ideal for evaluating the generalization of point cloud completion methods in complex urban environments.
To evaluate performance under extreme sparsity, noisiness and non-uniform sampling, we extend the Building-PCC dataset by generating two simulated point cloud datasets, aligned with the setup in ArcPro~\cite{huang2025arcpro}: 1) SfM scenario, which simulates point clouds obtained from multi-view image-based reconstruction. It employs perceptually-weighted sampling, favoring salient regions like sharp edges and upward-facing roofs to mimic non-uniform SfM feature matching. The simulation further incorporates Gaussian noise to represent re-projection uncertainty and deliberately injects spatial outliers, closely replicating the sparse and noisy characteristics of real-world SfM data.
%
%
2) Sparse scenario, created by progressively removing neighborhood points through a random anchor-driven strategy, simulating severe data loss, with uniform or Perlin noise added for realism. The number of points for each point cloud is randomly and uniformly sampled from the range of 200 to 2000. This design explicitly controls the level of sparsity, ensuring that the model is evaluated across a continuous and challenging spectrum of data completeness.


\begin{figure*}[th!]
  \centering
   \includegraphics[width=1\linewidth]{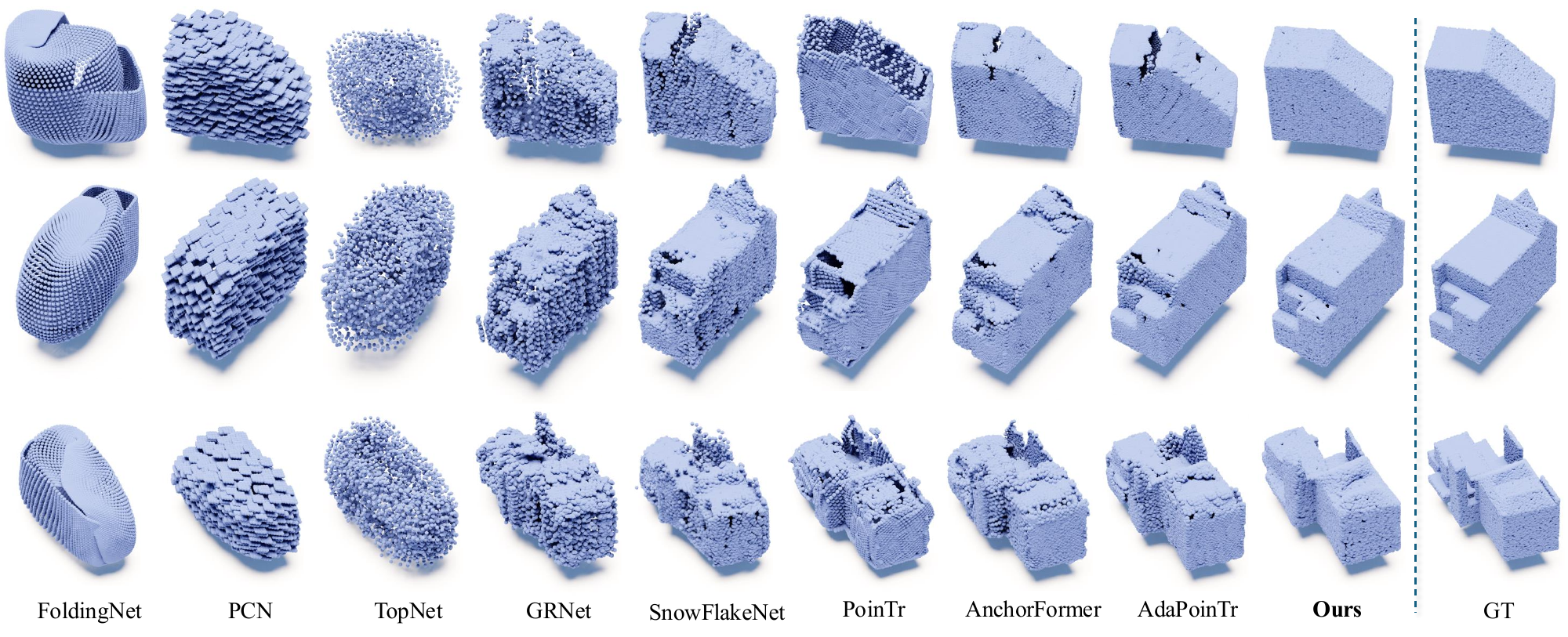}
    \vspace{-8mm}
   \caption{\textbf{3D Building Point Cloud Completion Comparison.} Our method is compared qualitatively against several established baselines known for their strong performance on building point cloud completion~\cite{isprs-annals-X-4-W5-2024-179-2024}. Our generative 3D priors faithfully capture intricate geometries and thin structures, and consistently produce more complete and uniform results.}
   \label{fig:main-pcc-result}
   \vspace{-3mm}
\end{figure*}

\begin{table}[t!]
\caption{\textbf{Quantitative comparison on structured mesh abstraction.} Our method achieves the best performance across all metrics, calculated as the average of three scenarios. Refer to the \textit{Supplementary Material} for the full set of experimental records.}
\vspace{-2mm}
\label{tab:mesh_main}
\centering
\small
\setlength{\tabcolsep}{1.5mm}{
\scalebox{1}{
\begin{tabular}{l|ccccc}
\toprule
 Methods & \# V $\downarrow$ & \# F $\downarrow$ & \# P $\downarrow$ & FR $\downarrow$ &CD $\downarrow$\\
\midrule
City3D~\cite{huang2022city3d}               & 173 & 72 & 14 & 6 \% & 0.167 \\
Point2Building~\cite{liu2024point2building} & 20 & 34 & 18 & 1 \% & 0.043\\
\midrule
\textbf{Ours} & \textbf{10} & \textbf{16} & \textbf{8} & \textbf{0 \%} & \textbf{0.036}\\
\bottomrule
\end{tabular}}
}
\captionsetup{justification=justified}
\vspace{-6mm}
\end{table}

\noindent\textbf{Metrics.}
To rigorously evaluate our reconstructed point clouds for artistic mesh abstraction, we employ four complementary metrics measuring geometric fidelity, distribution consistency, and point uniformity, properties that are all critical for high-quality mesh generation. Unlike tasks focused solely on completeness, which have been shown to be insufficient for polygonal surface reconstruction~\cite{chen2025parametric}, our approach requires point clouds that form a clean and well-distributed suggestive surface.
We utilize Chamfer Distance (CD) to assess overall shape accuracy, while Earth Mover's Distance (EMD) complements CD by capturing global distribution consistency and surface continuity, aspects to which CD is inherently insensitive. Beyond gross geometry, F-score evaluates the recovery of fine details and sharp features at a predefined distance threshold, reflecting perceptual structural quality. Finally, a uniformity metric quantifies distribution regularity~\cite{Li_2019_ICCV}, ensuring the point cloud supports the creation of a clean and artistically viable mesh.
%
%
For the generated meshes, we evaluate their geometric simplicity and structural consistency using standard metrics: vertex count (\#V), face count (\#F), and plane count (\#P). These metrics are typically adopted in artistic mesh generation and structural abstraction~\cite{siddiqui2024meshgpt,huang2025arcpro}. 
Geometric accuracy is assessed primarily by CD, complemented by the Failure Rate (FR) for a comprehensive evaluation.
Please refer to \textit{Supplementary Material} for comprehensive experimental settings and detailed metric calculations.



\noindent\textbf{Baselines.}
Based on the comprehensive benchmark established in Building-PCC~\cite{isprs-annals-X-4-W5-2024-179-2024}, we compare our method against eight established and highly-cited baselines of point cloud completion methods to ensure a fair and thorough comparison. These methods cover a diverse range of deep learning strategies, including encoder-decoder architectures (PCN~\cite{yuan2018pcn}, FoldingNet~\cite{yang2018foldingnet}), tree-structured decoders (TopNet~\cite{tchapmi2019topnet}), grid-based regularization (GRNet~\cite{xie2020grnet}), detail-oriented deconvolution (SnowflakeNet~\cite{xiang2021snowflakenet}), and transformer-based frameworks (PoinTr~\cite{yu2021pointr}, AnchorFormer~\cite{chen2023anchorformer}, AdaPoinTr~\cite{yu2023adapointr}).
All baseline models are evaluated using their publicly available pre-trained weights from the Building-PCC benchmark. 
For the generated meshes, we evaluated them against the two open-source 3D building structural abstraction methods, which represent the state-of-the-art performance in both learning-based~\cite{liu2024point2building} and traditional solver-based~\cite{huang2022city3d} approaches.

%


\subsection{Building Structured Abstraction Comparison}
Our method demonstrates superior quantitative~\cref{tab:mesh_main} and qualitative~\cref{fig:main-result} performance across all scenarios. As shown~\cref{fig:main-result}, City3D (a robust variant of PolyFit~\cite{nan2017polyfit}) performs adequately on airborne LiDAR point clouds, for which it was specifically designed, but fails to generalize to other scenarios, producing shapes that deviate significantly from the target distributions. The point-conditioned autoregressive transformer in Point2Building succeeds in recovering coarse structural outlines, particularly in the LiDAR scenario—a fact reflected in its CD scores (0.043, which is only marginally higher than our score of 0.036 \textcolor{green}{$\downarrow$} in~\cref{tab:mesh_main}). However, its decoupled modeling of vertices and faces leads to severe inconsistencies during mesh inference, resulting in visually chaotic artifacts and an inability to reconstruct more complex geometries (failing in the SfM scenario). In contrast, our method robustly reconstructs low-poly meshes that are highly consistent with manual annotations. This is made possible by the recovered intermediate point clouds, which provide a critical structural guide.

\begin{table}[t!]
\caption{\textbf{Quantitative comparison on point cloud completion.} Our method achieves the best performance across all metrics, demonstrating superior completeness (F-score), geometric accuracy (CD, EMD), and point distribution uniformity, measured for LiDAR point cloud scenarios.}
\vspace{-2mm}
\label{tab:pc_main}
\centering
\small
\setlength{\tabcolsep}{0.4mm}{
\scalebox{1}{
\begin{tabular}{l|cccc}
\toprule
 Methods & F-score$\uparrow$ & CD $\downarrow$ & Uniformity $\downarrow$ & EMD$\downarrow$ \\
\midrule
FoldingNet~\cite{yang2018foldingnet}        & 0.41 & 1.47 & 2.43 & 0.29 \\
PCN~\cite{yuan2018pcn}                      & 0.86 & 0.74 & 0.69 & 0.15 \\
TopNet~\cite{tchapmi2019topnet}             & 0.25 & 0.97 & 0.61 & 0.35 \\
GRNet~\cite{xie2020grnet}                   & 0.74 & 1.41 & 1.13 & 0.19 \\
SnowFlakeNet~\cite{xiang2021snowflakenet}   & 0.60 & 1.38 & 1.52 & 0.17 \\
PoinTr~\cite{yu2021pointr}                  & 0.85 & 0.41 & 0.25 & 0.12 \\
AnchorFormer~\cite{chen2023anchorformer}    & 0.82 & 0.39 & 1.27 & 0.13 \\
AdaPoinTr~\cite{yu2023adapointr}            & 0.70 & 0.44 & 1.49 & 0.15 \\
\midrule
\textbf{Ours} & \textbf{0.91} & \textbf{0.35} & \textbf{0.04} & \textbf{0.10} \\
\bottomrule
\end{tabular}}
}
\captionsetup{justification=justified}
\vspace{-6mm}
\end{table}

\subsection{Building Point Cloud Completion Comparison}
We further demonstrate that our recovered point cloud distribution, which serves solely as an intermediate representation within our generative framework, yields the most visually compelling results, as shown in~\cref{fig:main-pcc-result}. This is evident in terms of surface proximity, 3D reconstruction fidelity (particularly for intricate structures), and distribution uniformity. The superior quality is also quantitatively confirmed in~\cref{tab:pc_main}, where our method achieves a uniformity score nearly an order of magnitude lower than all competing approaches (0.04 \textcolor{green}{$\downarrow$}), while outperforming our baselines across all other standard accuracy metrics.

\begin{figure}[t!]
\centering
\includegraphics[width=0.9\linewidth]{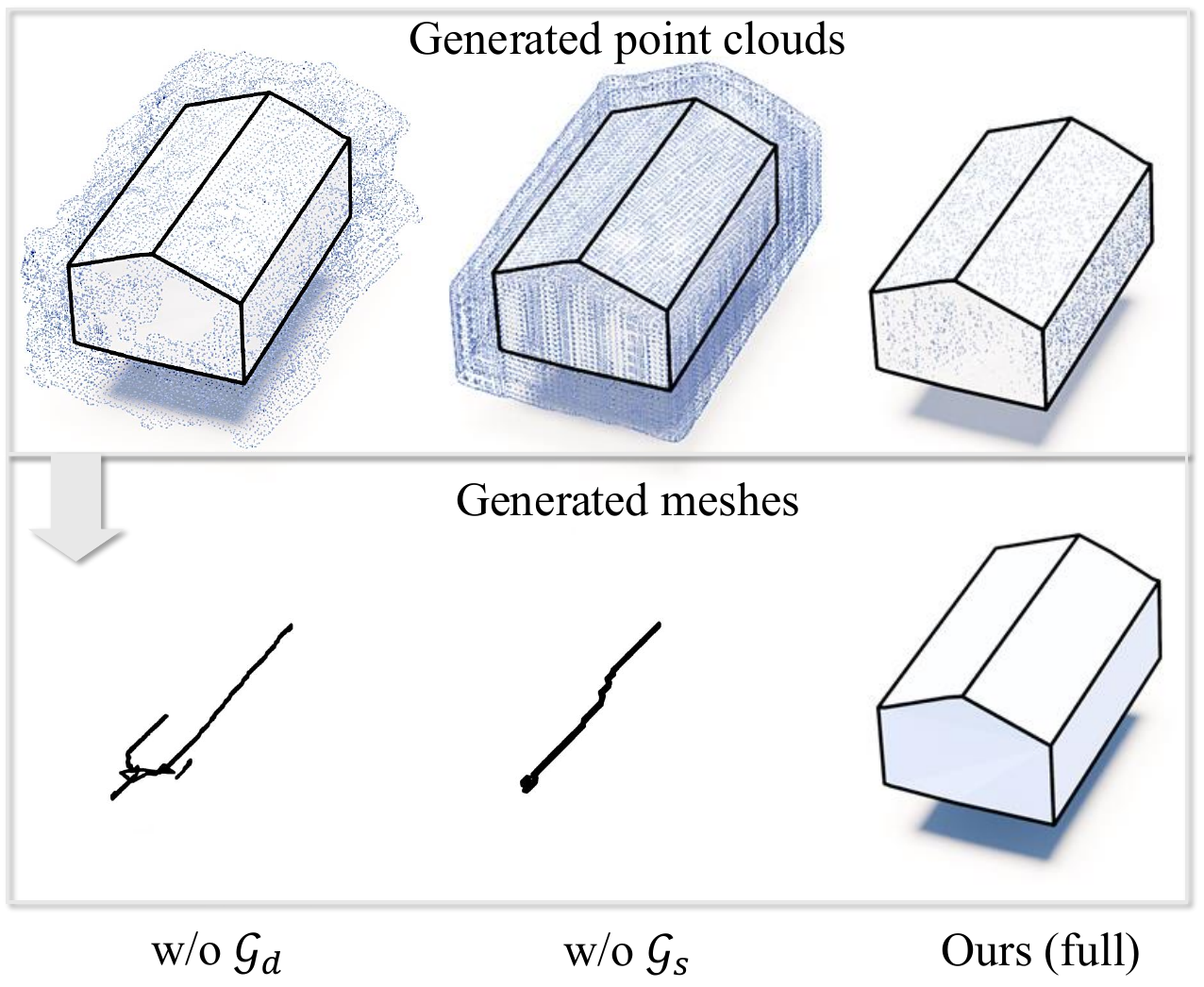}
\caption{\textbf{Ablation on Cascaded Designs in Loca-DiT: Diffusion.} When critical modules are removed (top row, left and middle), the intermediate point clouds deviate significantly from the target surface (mesh). Our full model (top row, right) corrects this, producing a point distribution that faithfully captures the target geometry. The bottom row shows the final meshes generated autoregressively from these point clouds.}
\label{fig:ablate-diff}
\vspace{-6mm}
\end{figure}

\subsection{Ablation Study}
\begin{figure}[t!]
\centering
\includegraphics[width=1\linewidth]{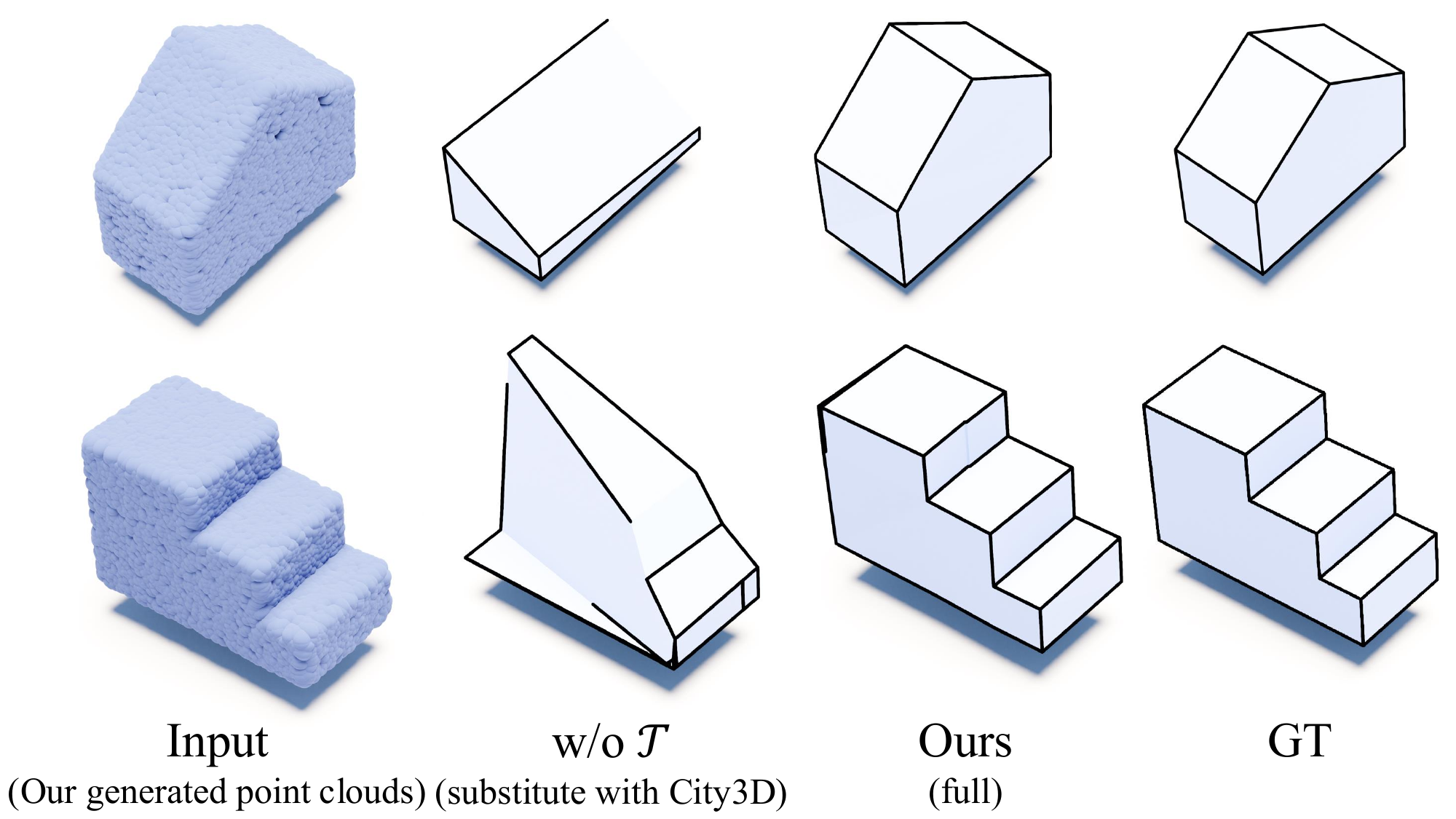}
\vspace{-4mm}
\caption{\textbf{Ablation on Cascaded Designs in Loca-DiT: Transformer.} While provided with the same intermediate point cloud, the traditional solver~\cite{huang2022city3d} still cannot generate a valid surface, unlike our autoregressive transformer.}
\label{fig:ablate-trans}
\vspace{-2mm}
\end{figure}

The core of Loca-DiT lies in a series of tightly coupled latent space transitions, which progressively transform the embedding from a state favoring unstructured point clouds to one suitable for structured representations—commonly encoded as sequential tokens in recent works.
We conduct an ablation study on each module governing a specific part of this latent formulation: namely \(\mathcal{G}_{d}\), \(\mathcal{G}_{s}\), and \(\mathcal{T}\). As shown in~\cref{fig:ablate-diff}, removing the coarse-level latent \(\mathcal{G}_{d}\) leads to chaotic point clouds that fail to reflect the target shape. Without the fine-level grid latent \(\mathcal{G}_{s}\), the model can only recover preliminary outlines; however, since the point cloud is retrieved from grid vertices, this results in a \textquotedblleft double-surface effect \textquotedblright that misleads subsequent mesh generation. Only when both grid latents are incorporated does the recovered point cloud closely conform to the mesh surfaces.
We further demonstrate that our tokenization scheme is indispensable, as replacing the entire transformer with a traditional solver (using our recovered distribution as input) proves infeasible (see~\cref{fig:ablate-trans}). 
While these results affirm the generalizability of modern autoregressive transformers, their performance is contingent upon our key innovation: the incorporation of 3D generative priors. This is confirmed in~\cref{tab:ablate-prior}, which shows a significant decline in mesh quality when the prior, embodied by 
\(\mathcal{P}_{out}\), is removed. Both settings are evaluated using MeshAnything V2, which is pre-trained on ground-truth building mesh-point cloud pairs to learn the data distribution. The trained model is then evaluated directly on both \(\mathcal{P}_{in}\) and \(\mathcal{P}_{out}\).

\section{Discussion}
\noindent\textbf{Limitations.} 
A persistent bias in publicly available building datasets toward simple geometries inherently limits the modeling of complex details. As a result, despite our method's substantial gains over the baselines, its performance on intricate structures remains constrained.

\noindent\textbf{Future works.} 
The modularity of our framework underpins its future-proof scalability, as the 3D diffusion and autoregressive components can be upgraded independently with advancements in each subfield. To further enhance performance, especially in precision-sensitive applications, future work could focus on integrating stronger priors, such as height and geographic coordinate embeddings. Ultimately, sustained progress in this domain will rely on the community-driven development of building datasets that feature greater geometric diversity and incorporate raw observations from a broader range of sources.


\begin{table}[t!]
\caption{\textbf{Ablation on 3D generative priors in autoregressive Transformer.} We test MeshAnything V2 on both the $\mathcal{P}_{in}$ (w/o priors) and $\mathcal{P}_{out}$ (ours).}
\vspace{-2mm}
\label{tab:ablate-prior}
\centering
\small
\setlength{\tabcolsep}{1.5mm}{
\scalebox{1}{
\begin{tabular}{l|ccccc}
\toprule
 Settings & \# V $\downarrow$ & \# F $\downarrow$ & \# P $\downarrow$ & FR $\downarrow$ &CD $\downarrow$\\
\midrule
w/o priors             & 78 & 127 & 53 & \textless 1 \% & 0.107 \\
\textbf{Ours} & \textbf{38} & \textbf{70} & \textbf{23} & \textbf{0 \%} & \textbf{0.034}\\
\bottomrule
\end{tabular}}
}
\captionsetup{justification=justified}
\vspace{-4mm}
\end{table}

\section{Conclusion}
We introduced BuildAnyPoint, a generative framework for structured 3D building reconstruction from point clouds of arbitrary distributions. By leveraging a Loosely Cascaded Diffusion Transformer, termed Loca-DiT, our approach explicitly recovers underlying geometric priors and autoregressively generates high-quality meshes. Experiments demonstrate state-of-the-art performance in both building abstraction and point cloud completion, with strong generalization across diverse urban data sources. 

{
    \small
    \bibliographystyle{ieeenat_fullname}
    \bibliography{main}

\begin{thebibliography}{61}
\providecommand{\natexlab}[1]{#1}
\providecommand{\url}[1]{\texttt{#1}}
\expandafter\ifx\csname urlstyle\endcsname\relax
  \providecommand{\doi}[1]{doi: #1}\else
  \providecommand{\doi}{doi: \begingroup \urlstyle{rm}\Url}\fi

\bibitem[Alliegro et~al.(2023)Alliegro, Siddiqui, Tommasi, and
  Nie{\ss}ner]{alliegro2023polydiff}
Antonio Alliegro, Yawar Siddiqui, Tatiana Tommasi, and Matthias Nie{\ss}ner.
\newblock Polydiff: Generating 3d polygonal meshes with diffusion models.
\newblock \emph{arXiv preprint arXiv:2312.11417}, 2023.

\bibitem[Bauchet and Lafarge(2019)]{buildbauchet2019city}
J-P Bauchet and Florent Lafarge.
\newblock City reconstruction from airborne lidar: A computational geometry
  approach.
\newblock \emph{ISPRS Annals of the Photogrammetry, Remote Sensing and Spatial
  Information Sciences}, 4:\penalty0 19--26, 2019.

\bibitem[Bauchet and Lafarge(2020)]{bauchet2020kinetic}
Jean-Philippe Bauchet and Florent Lafarge.
\newblock Kinetic shape reconstruction.
\newblock \emph{ACM Transactions on Graphics (TOG)}, 39\penalty0 (5):\penalty0
  1--14, 2020.

\bibitem[Bauchet et~al.(2024{\natexlab{a}})Bauchet, Sulzer, Lafarge, and
  Tarabalka]{bauchet2024simplicity}
Jean-Philippe Bauchet, Raphael Sulzer, Florent Lafarge, and Yuliya Tarabalka.
\newblock Simplicity: Reconstructing buildings with simple regularized 3d
  models.
\newblock In \emph{Proceedings of the IEEE/CVF Conference on Computer Vision
  and Pattern Recognition}, pages 7616--7626, 2024{\natexlab{a}}.

\bibitem[Bauchet et~al.(2024{\natexlab{b}})Bauchet, Sulzer, Lafarge, and
  Tarabalka]{buildbauchet2024simplicity}
Jean-Philippe Bauchet, Raphael Sulzer, Florent Lafarge, and Yuliya Tarabalka.
\newblock Simplicity: Reconstructing buildings with simple regularized 3d
  models.
\newblock In \emph{Proceedings of the IEEE/CVF Conference on Computer Vision
  and Pattern Recognition}, pages 7616--7626, 2024{\natexlab{b}}.

\bibitem[Biljecki et~al.(2015)Biljecki, Stoter, Ledoux, Zlatanova, and
  {\c{C}}{\"o}ltekin]{biljecki2015applications}
Filip Biljecki, Jantien Stoter, Hugo Ledoux, Sisi Zlatanova, and Arzu
  {\c{C}}{\"o}ltekin.
\newblock Applications of 3d city models: State of the art review.
\newblock \emph{ISPRS International Journal of Geo-Information}, 4\penalty0
  (4):\penalty0 2842--2889, 2015.

\bibitem[Chen et~al.(2024{\natexlab{a}})Chen, Chen, Pang, Zeng, Cheng, Fu, Yin,
  Wang, Yu, Yu, et~al.]{chen2024meshxl}
Sijin Chen, Xin Chen, Anqi Pang, Xianfang Zeng, Wei Cheng, Yijun Fu, Fukun Yin,
  Billzb Wang, Jingyi Yu, Gang Yu, et~al.
\newblock Meshxl: Neural coordinate field for generative 3d foundation models.
\newblock \emph{Advances in Neural Information Processing Systems},
  37:\penalty0 97141--97166, 2024{\natexlab{a}}.

\bibitem[Chen et~al.(2024{\natexlab{b}})Chen, He, Huang, Ye, Chen, Tang, Chen,
  Cai, Yang, Yu, et~al.]{chen2024meshanything}
Yiwen Chen, Tong He, Di Huang, Weicai Ye, Sijin Chen, Jiaxiang Tang, Xin Chen,
  Zhongang Cai, Lei Yang, Gang Yu, et~al.
\newblock Meshanything: Artist-created mesh generation with autoregressive
  transformers.
\newblock \emph{arXiv preprint arXiv:2406.10163}, 2024{\natexlab{b}}.

\bibitem[Chen et~al.(2024{\natexlab{c}})Chen, Wang, Luo, Wang, Chen, Zhu,
  Zhang, and Lin]{chen2024meshanythingv2}
Yiwen Chen, Yikai Wang, Yihao Luo, Zhengyi Wang, Zilong Chen, Jun Zhu, Chi
  Zhang, and Guosheng Lin.
\newblock Meshanything v2: Artist-created mesh generation with adjacent mesh
  tokenization.
\newblock \emph{arXiv preprint arXiv:2408.02555}, 2024{\natexlab{c}}.

\bibitem[Chen et~al.(2022)Chen, Ledoux, Khademi, and
  Nan]{chen2022reconstructing}
Zhaiyu Chen, Hugo Ledoux, Seyran Khademi, and Liangliang Nan.
\newblock Reconstructing compact building models from point clouds using deep
  implicit fields.
\newblock \emph{ISPRS Journal of Photogrammetry and Remote Sensing},
  194:\penalty0 58--73, 2022.

\bibitem[Chen et~al.(2023)Chen, Long, Qiu, Yao, Zhou, Luo, and
  Mei]{chen2023anchorformer}
Zhikai Chen, Fuchen Long, Zhaofan Qiu, Ting Yao, Wengang Zhou, Jiebo Luo, and
  Tao Mei.
\newblock Anchorformer: Point cloud completion from discriminative nodes.
\newblock In \emph{Proceedings of the IEEE/CVF conference on computer vision
  and pattern recognition}, pages 13581--13590, 2023.

\bibitem[Chen et~al.(2024{\natexlab{d}})Chen, Shi, Nan, Xiong, and
  Zhu]{chen2024polygnn}
Zhaiyu Chen, Yilei Shi, Liangliang Nan, Zhitong Xiong, and Xiao~Xiang Zhu.
\newblock Polygnn: Polyhedron-based graph neural network for 3d building
  reconstruction from point clouds.
\newblock \emph{ISPRS Journal of Photogrammetry and Remote Sensing},
  218:\penalty0 693--706, 2024{\natexlab{d}}.

\bibitem[Chen et~al.(2025)Chen, Wang, Nan, and Zhu]{chen2025parametric}
Zhaiyu Chen, Yuqing Wang, Liangliang Nan, and Xiao~Xiang Zhu.
\newblock Parametric point cloud completion for polygonal surface
  reconstruction.
\newblock In \emph{Proceedings of the Computer Vision and Pattern Recognition
  Conference}, pages 11749--11758, 2025.

\bibitem[Du et~al.(2024)Du, Yan, Wang, Xie, and Pu]{du2024arbitrary}
Hang Du, Xuejun Yan, Jingjing Wang, Di Xie, and Shiliang Pu.
\newblock Arbitrary-scale point cloud upsampling by voxel-based network with
  latent geometric-consistent learning.
\newblock In \emph{Proceedings of the AAAI conference on artificial
  intelligence}, pages 1626--1634, 2024.

\bibitem[Feng et~al.(2022)Feng, Li, Cai, Luo, and Zhang]{Feng_2022_CVPR}
Wanquan Feng, Jin Li, Hongrui Cai, Xiaonan Luo, and Juyong Zhang.
\newblock Neural points: Point cloud representation with neural fields for
  arbitrary upsampling.
\newblock In \emph{Proceedings of the IEEE/CVF Conference on Computer Vision
  and Pattern Recognition (CVPR)}, pages 18633--18642, 2022.

\bibitem[Gao et~al.(2024)Gao, Peters, and
  Stoter]{isprs-annals-X-4-W5-2024-179-2024}
W. Gao, R. Peters, and J. Stoter.
\newblock Building-pcc: Building point cloud completion benchmarks.
\newblock \emph{ISPRS Annals of the Photogrammetry, Remote Sensing and Spatial
  Information Sciences}, X-4/W5-2024:\penalty0 179--186, 2024.

\bibitem[Gupta et~al.(2023)Gupta, Xiong, Nie, Jones, and
  O{\u{g}}uz]{nigupta20233dgen}
Anchit Gupta, Wenhan Xiong, Yixin Nie, Ian Jones, and Barlas O{\u{g}}uz.
\newblock 3dgen: Triplane latent diffusion for textured mesh generation.
\newblock \emph{arXiv preprint arXiv:2303.05371}, 2023.

\bibitem[Hao et~al.(2024)Hao, Romero, Lin, and Liu]{hao2024meshtron}
Zekun Hao, David~W Romero, Tsung-Yi Lin, and Ming-Yu Liu.
\newblock Meshtron: High-fidelity, artist-like 3d mesh generation at scale.
\newblock \emph{arXiv preprint arXiv:2412.09548}, 2024.

\bibitem[He et~al.(2025)He, Kwon, Cai, and Adeli]{he2025artist}
Yao He, Youngjoong Kwon, Wenxiao Cai, and Ehsan Adeli.
\newblock Artist-created mesh generation from raw observation.
\newblock In \emph{Proceedings of the IEEE/CVF International Conference on
  Computer Vision}, pages 2642--2647, 2025.

\bibitem[Hua et~al.(2025)Hua, Jiang, Chen, and Zhao]{hua2025sat2city}
Tongyan Hua, Lutao Jiang, Ying-Cong Chen, and Wufan Zhao.
\newblock Sat2city: 3d city generation from a single satellite image with
  cascaded latent diffusion.
\newblock In \emph{Proceedings of the IEEE/CVF International Conference on
  Computer Vision}, pages 27978--27988, 2025.

\bibitem[Huang et~al.(2022{\natexlab{a}})Huang, Stoter, Peters, and
  Nan]{buildhuang2022city3d}
Jin Huang, Jantien Stoter, Ravi Peters, and Liangliang Nan.
\newblock City3d: Large-scale building reconstruction from airborne lidar point
  clouds.
\newblock \emph{Remote Sensing}, 14\penalty0 (9):\penalty0 2254,
  2022{\natexlab{a}}.

\bibitem[Huang et~al.(2022{\natexlab{b}})Huang, Stoter, Peters, and
  Nan]{huang2022city3d}
Jin Huang, Jantien Stoter, Ravi Peters, and Liangliang Nan.
\newblock City3d: Large-scale building reconstruction from airborne lidar point
  clouds.
\newblock \emph{Remote Sensing}, 14\penalty0 (9):\penalty0 2254,
  2022{\natexlab{b}}.

\bibitem[Huang et~al.(2025)Huang, Zhang, Liu, Gong, Zhang, and
  Huang]{huang2025arcpro}
Qirui Huang, Runze Zhang, Kangjun Liu, Minglun Gong, Hao Zhang, and Hui Huang.
\newblock Arcpro: Architectural programs for structured 3d abstraction of
  sparse points.
\newblock In \emph{Proceedings of the Computer Vision and Pattern Recognition
  Conference}, pages 6563--6572, 2025.

\bibitem[Hui et~al.(2022)Hui, Li, Hu, and Fu]{gridhui2022neural}
Ka-Hei Hui, Ruihui Li, Jingyu Hu, and Chi-Wing Fu.
\newblock Neural wavelet-domain diffusion for 3d shape generation.
\newblock In \emph{SIGGRAPH Asia 2022 conference papers}, pages 1--9, 2022.

\bibitem[Li et~al.(2022)Li, Song, Sun, Liu, Wang, Yao, and
  Cao]{li2022point2roof}
Li Li, Nan Song, Fei Sun, Xinyi Liu, Ruisheng Wang, Jian Yao, and Shaosheng
  Cao.
\newblock Point2roof: End-to-end 3d building roof modeling from airborne lidar
  point clouds.
\newblock \emph{ISPRS Journal of Photogrammetry and Remote Sensing},
  193:\penalty0 17--28, 2022.

\bibitem[Li et~al.(2019)Li, Li, Fu, Cohen-Or, and Heng]{Li_2019_ICCV}
Ruihui Li, Xianzhi Li, Chi-Wing Fu, Daniel Cohen-Or, and Pheng-Ann Heng.
\newblock Pu-gan: A point cloud upsampling adversarial network.
\newblock In \emph{Proceedings of the IEEE/CVF International Conference on
  Computer Vision (ICCV)}, 2019.

\bibitem[Lin et~al.(2023)Lin, Gao, Tang, Takikawa, Zeng, Huang, Kreis, Fidler,
  Liu, and Lin]{2dlin2023magic3d}
Chen-Hsuan Lin, Jun Gao, Luming Tang, Towaki Takikawa, Xiaohui Zeng, Xun Huang,
  Karsten Kreis, Sanja Fidler, Ming-Yu Liu, and Tsung-Yi Lin.
\newblock Magic3d: High-resolution text-to-3d content creation.
\newblock In \emph{Proceedings of the IEEE/CVF conference on computer vision
  and pattern recognition}, pages 300--309, 2023.

\bibitem[Liu et~al.(2023)Liu, Xu, Jin, Chen, Varma~T, Xu, and Su]{niliu2023one}
Minghua Liu, Chao Xu, Haian Jin, Linghao Chen, Mukund Varma~T, Zexiang Xu, and
  Hao Su.
\newblock One-2-3-45: Any single image to 3d mesh in 45 seconds without
  per-shape optimization.
\newblock \emph{Advances in Neural Information Processing Systems},
  36:\penalty0 22226--22246, 2023.

\bibitem[Liu et~al.(2024)Liu, Obukhov, Wegner, and
  Schindler]{liu2024point2building}
Yujia Liu, Anton Obukhov, Jan~Dirk Wegner, and Konrad Schindler.
\newblock Point2building: Reconstructing buildings from airborne lidar point
  clouds.
\newblock \emph{ISPRS Journal of Photogrammetry and Remote Sensing},
  215:\penalty0 351--368, 2024.

\bibitem[Luo and Hu(2021)]{pcluo2021diffusion}
Shitong Luo and Wei Hu.
\newblock Diffusion probabilistic models for 3d point cloud generation.
\newblock In \emph{Proceedings of the IEEE/CVF conference on computer vision
  and pattern recognition}, pages 2837--2845, 2021.

\bibitem[Nagasawa et~al.(2021)Nagasawa, Mas, Moya, and
  Koshimura]{nagasawa2021model}
Ryosuke Nagasawa, Erick Mas, Luis Moya, and Shunichi Koshimura.
\newblock Model-based analysis of multi-uav path planning for surveying
  postdisaster building damage.
\newblock \emph{Scientific reports}, 11\penalty0 (1):\penalty0 18588, 2021.

\bibitem[Nan and Wonka(2017)]{nan2017polyfit}
Liangliang Nan and Peter Wonka.
\newblock Polyfit: Polygonal surface reconstruction from point clouds.
\newblock In \emph{Proceedings of the IEEE international conference on computer
  vision}, pages 2353--2361, 2017.

\bibitem[Nash et~al.(2020)Nash, Ganin, Eslami, and Battaglia]{nash2020polygen}
Charlie Nash, Yaroslav Ganin, SM~Ali Eslami, and Peter Battaglia.
\newblock Polygen: An autoregressive generative model of 3d meshes.
\newblock In \emph{International conference on machine learning}, pages
  7220--7229. PMLR, 2020.

\bibitem[Nichol et~al.(2022)Nichol, Jun, Dhariwal, Mishkin, and
  Chen]{pcnichol2022point}
Alex Nichol, Heewoo Jun, Prafulla Dhariwal, Pamela Mishkin, and Mark Chen.
\newblock Point-e: A system for generating 3d point clouds from complex
  prompts.
\newblock \emph{arXiv preprint arXiv:2212.08751}, 2022.

\bibitem[Pan et~al.(2025)Pan, Zhang, Liu, Gong, and Huang]{pan2025building}
Shanshan Pan, Runze Zhang, Yilin Liu, Minglun Gong, and Hui Huang.
\newblock Building lod representation for 3d urban scenes.
\newblock \emph{ISPRS Journal of Photogrammetry and Remote Sensing},
  226:\penalty0 16--32, 2025.

\bibitem[Poole et~al.(2022)Poole, Jain, Barron, and
  Mildenhall]{2dpoole2022dreamfusion}
Ben Poole, Ajay Jain, Jonathan~T Barron, and Ben Mildenhall.
\newblock Dreamfusion: Text-to-3d using 2d diffusion.
\newblock \emph{arXiv preprint arXiv:2209.14988}, 2022.

\bibitem[Ren et~al.(2024)Ren, Huang, Zeng, Museth, Fidler, and
  Williams]{ren2024xcube}
Xuanchi Ren, Jiahui Huang, Xiaohui Zeng, Ken Museth, Sanja Fidler, and Francis
  Williams.
\newblock Xcube: Large-scale 3d generative modeling using sparse voxel
  hierarchies.
\newblock In \emph{Proceedings of the IEEE/CVF Conference on Computer Vision
  and Pattern Recognition}, pages 4209--4219, 2024.

\bibitem[Siddiqui et~al.(2024)Siddiqui, Alliegro, Artemov, Tommasi, Sirigatti,
  Rosov, Dai, and Nie{\ss}ner]{siddiqui2024meshgpt}
Yawar Siddiqui, Antonio Alliegro, Alexey Artemov, Tatiana Tommasi, Daniele
  Sirigatti, Vladislav Rosov, Angela Dai, and Matthias Nie{\ss}ner.
\newblock Meshgpt: Generating triangle meshes with decoder-only transformers.
\newblock In \emph{Proceedings of the IEEE/CVF conference on computer vision
  and pattern recognition}, pages 19615--19625, 2024.

\bibitem[Tang et~al.(2023)Tang, Gu, Wang, Zhang, Bao, Chen, and
  Guo]{gridtang2023volumediffusion}
Zhicong Tang, Shuyang Gu, Chunyu Wang, Ting Zhang, Jianmin Bao, Dong Chen, and
  Baining Guo.
\newblock Volumediffusion: Flexible text-to-3d generation with efficient
  volumetric encoder.
\newblock \emph{arXiv preprint arXiv:2312.11459}, 2023.

\bibitem[Tchapmi et~al.(2019)Tchapmi, Kosaraju, Rezatofighi, Reid, and
  Savarese]{tchapmi2019topnet}
Lyne~P Tchapmi, Vineet Kosaraju, Hamid Rezatofighi, Ian Reid, and Silvio
  Savarese.
\newblock Topnet: Structural point cloud decoder.
\newblock In \emph{Proceedings of the IEEE/CVF conference on computer vision
  and pattern recognition}, pages 383--392, 2019.

\bibitem[Verdie et~al.(2015{\natexlab{a}})Verdie, Lafarge, and
  Alliez]{buildverdie2015lod}
Yannick Verdie, Florent Lafarge, and Pierre Alliez.
\newblock Lod generation for urban scenes.
\newblock \emph{ACM Transactions on Graphics}, 34\penalty0 (3):\penalty0 15,
  2015{\natexlab{a}}.

\bibitem[Verdie et~al.(2015{\natexlab{b}})Verdie, Lafarge, and
  Alliez]{verdie2015lod}
Yannick Verdie, Florent Lafarge, and Pierre Alliez.
\newblock Lod generation for urban scenes.
\newblock \emph{ACM Transactions on Graphics}, 34\penalty0 (3):\penalty0 15,
  2015{\natexlab{b}}.

\bibitem[Wang et~al.(2023{\natexlab{a}})Wang, Huang, and
  Yang]{wang2023building3d}
Ruisheng Wang, Shangfeng Huang, and Hongxin Yang.
\newblock Building3d: A urban-scale dataset and benchmarks for learning roof
  structures from point clouds.
\newblock In \emph{Proceedings of the IEEE/CVF International Conference on
  Computer Vision}, pages 20076--20086, 2023{\natexlab{a}}.

\bibitem[Wang et~al.(2023{\natexlab{b}})Wang, Zhang, Zhang, Gu, Bao,
  Baltrusaitis, Shen, Chen, Wen, Chen, et~al.]{niwang2023rodin}
Tengfei Wang, Bo Zhang, Ting Zhang, Shuyang Gu, Jianmin Bao, Tadas
  Baltrusaitis, Jingjing Shen, Dong Chen, Fang Wen, Qifeng Chen, et~al.
\newblock Rodin: A generative model for sculpting 3d digital avatars using
  diffusion.
\newblock In \emph{Proceedings of the IEEE/CVF conference on computer vision
  and pattern recognition}, pages 4563--4573, 2023{\natexlab{b}}.

\bibitem[Wang et~al.(2023{\natexlab{c}})Wang, Lu, Wang, Bao, Li, Su, and
  Zhu]{2dwang2023prolificdreamer}
Zhengyi Wang, Cheng Lu, Yikai Wang, Fan Bao, Chongxuan Li, Hang Su, and Jun
  Zhu.
\newblock Prolificdreamer: High-fidelity and diverse text-to-3d generation with
  variational score distillation.
\newblock \emph{Advances in neural information processing systems},
  36:\penalty0 8406--8441, 2023{\natexlab{c}}.

\bibitem[Weng et~al.(2024)Weng, Wang, Zhang, Chen, and Zhu]{weng2024pivotmesh}
Haohan Weng, Yikai Wang, Tong Zhang, CL Chen, and Jun Zhu.
\newblock Pivotmesh: Generic 3d mesh generation via pivot vertices guidance.
\newblock \emph{arXiv preprint arXiv:2405.16890}, 2024.

\bibitem[Xiang et~al.(2025)Xiang, Lv, Xu, Deng, Wang, Zhang, Chen, Tong, and
  Yang]{3dxiang2025structured}
Jianfeng Xiang, Zelong Lv, Sicheng Xu, Yu Deng, Ruicheng Wang, Bowen Zhang,
  Dong Chen, Xin Tong, and Jiaolong Yang.
\newblock Structured 3d latents for scalable and versatile 3d generation.
\newblock In \emph{Proceedings of the Computer Vision and Pattern Recognition
  Conference}, pages 21469--21480, 2025.

\bibitem[Xiang et~al.(2021)Xiang, Wen, Liu, Cao, Wan, Zheng, and
  Han]{xiang2021snowflakenet}
Peng Xiang, Xin Wen, Yu-Shen Liu, Yan-Pei Cao, Pengfei Wan, Wen Zheng, and
  Zhizhong Han.
\newblock Snowflakenet: Point cloud completion by snowflake point deconvolution
  with skip-transformer.
\newblock In \emph{Proceedings of the IEEE/CVF international conference on
  computer vision}, pages 5499--5509, 2021.

\bibitem[Xie et~al.(2020)Xie, Yao, Zhou, Mao, Zhang, and Sun]{xie2020grnet}
Haozhe Xie, Hongxun Yao, Shangchen Zhou, Jiageng Mao, Shengping Zhang, and
  Wenxiu Sun.
\newblock Grnet: Gridding residual network for dense point cloud completion.
\newblock In \emph{European conference on computer vision}, pages 365--381.
  Springer, 2020.

\bibitem[Yang et~al.(2022)Yang, Cai, Du, Chen, Su, Wu, Wang, and
  Li]{yang2022connectivity}
Shengming Yang, Guorong Cai, Jing Du, Ping Chen, Jinhe Su, Yundong Wu, Zongyue
  Wang, and Jonathan Li.
\newblock Connectivity-aware graph: A planar topology for 3d building surface
  reconstruction.
\newblock \emph{ISPRS Journal of Photogrammetry and Remote Sensing},
  191:\penalty0 302--314, 2022.

\bibitem[Yang et~al.(2018)Yang, Feng, Shen, and Tian]{yang2018foldingnet}
Yaoqing Yang, Chen Feng, Yiru Shen, and Dong Tian.
\newblock Foldingnet: Point cloud auto-encoder via deep grid deformation.
\newblock In \emph{Proceedings of the IEEE conference on computer vision and
  pattern recognition}, pages 206--215, 2018.

\bibitem[Yu et~al.(2021)Yu, Rao, Wang, Liu, Lu, and Zhou]{yu2021pointr}
Xumin Yu, Yongming Rao, Ziyi Wang, Zuyan Liu, Jiwen Lu, and Jie Zhou.
\newblock Pointr: Diverse point cloud completion with geometry-aware
  transformers.
\newblock In \emph{Proceedings of the IEEE/CVF international conference on
  computer vision}, pages 12498--12507, 2021.

\bibitem[Yu et~al.(2023)Yu, Rao, Wang, Lu, and Zhou]{yu2023adapointr}
Xumin Yu, Yongming Rao, Ziyi Wang, Jiwen Lu, and Jie Zhou.
\newblock Adapointr: Diverse point cloud completion with adaptive
  geometry-aware transformers, 2023.

\bibitem[Yuan et~al.(2025)Yuan, Yang, Wang, Pan, Ma, Zhang, Liu, Cui, and
  Ma]{yuan2025immersegen}
Jinyan Yuan, Bangbang Yang, Keke Wang, Panwang Pan, Lin Ma, Xuehai Zhang, Xiao
  Liu, Zhaopeng Cui, and Yuewen Ma.
\newblock Immersegen: Agent-guided immersive world generation with
  alpha-textured proxies.
\newblock \emph{arXiv preprint arXiv:2506.14315}, 2025.

\bibitem[Yuan et~al.(2018)Yuan, Khot, Held, Mertz, and Hebert]{yuan2018pcn}
Wentao Yuan, Tejas Khot, David Held, Christoph Mertz, and Martial Hebert.
\newblock Pcn: Point completion network.
\newblock In \emph{2018 international conference on 3D vision (3DV)}, pages
  728--737. IEEE, 2018.

\bibitem[Zhang et~al.(2023)Zhang, Tang, Niessner, and
  Wonka]{3dzhang20233dshape2vecset}
Biao Zhang, Jiapeng Tang, Matthias Niessner, and Peter Wonka.
\newblock 3dshape2vecset: A 3d shape representation for neural fields and
  generative diffusion models.
\newblock \emph{ACM Transactions On Graphics (TOG)}, 42\penalty0 (4):\penalty0
  1--16, 2023.

\bibitem[Zhang et~al.(2024)Zhang, Wang, Zhang, Qiu, Pang, Jiang, Yang, Xu, and
  Yu]{3dzhang2024clay}
Longwen Zhang, Ziyu Wang, Qixuan Zhang, Qiwei Qiu, Anqi Pang, Haoran Jiang, Wei
  Yang, Lan Xu, and Jingyi Yu.
\newblock Clay: A controllable large-scale generative model for creating
  high-quality 3d assets.
\newblock \emph{ACM Transactions on Graphics (TOG)}, 43\penalty0 (4):\penalty0
  1--20, 2024.

\bibitem[Zhao et~al.(2023)Zhao, Liu, Chen, Zeng, Wang, Cheng, Fu, Chen, Yu, and
  Gao]{zhao2023michelangelo}
Zibo Zhao, Wen Liu, Xin Chen, Xianfang Zeng, Rui Wang, Pei Cheng, Bin Fu, Tao
  Chen, Gang Yu, and Shenghua Gao.
\newblock Michelangelo: Conditional 3d shape generation based on
  shape-image-text aligned latent representation.
\newblock \emph{Advances in neural information processing systems},
  36:\penalty0 73969--73982, 2023.

\bibitem[Zhao et~al.(2025)Zhao, Lai, Lin, Zhao, Liu, Yang, Feng, Yang, Zhang,
  Yang, et~al.]{zhao2025hunyuan3d}
Zibo Zhao, Zeqiang Lai, Qingxiang Lin, Yunfei Zhao, Haolin Liu, Shuhui Yang,
  Yifei Feng, Mingxin Yang, Sheng Zhang, Xianghui Yang, et~al.
\newblock Hunyuan3d 2.0: Scaling diffusion models for high resolution textured
  3d assets generation.
\newblock \emph{arXiv preprint arXiv:2501.12202}, 2025.

\bibitem[Zhou and Neumann(2012)]{buildzhou20122}
Qian-Yi Zhou and Ulrich Neumann.
\newblock 2.5 d building modeling by discovering global regularities.
\newblock In \emph{2012 IEEE Conference on Computer Vision and Pattern
  Recognition}, pages 326--333. IEEE, 2012.

\bibitem[Zhu et~al.(2024)Zhu, Yan, Wang, Zhang, Liu, and Zhang]{zhu2024lod}
Juelin Zhu, Shen Yan, Long Wang, Shengyue Zhang, Yu Liu, and Maojun Zhang.
\newblock Lod-loc: Aerial visual localization using lod 3d map with neural
  wireframe alignment.
\newblock \emph{Advances in Neural Information Processing Systems},
  37:\penalty0 119063--119098, 2024.

\end{thebibliography}
}

 \clearpage
\setcounter{page}{1}
\maketitlesupplementary

\section{More Technical details}

\subsection{Training Details}
\textbf{Diffusion models} are primarily implemented based on the sparse VAE and diffusion structure provided by XCube~\cite{ren2024xcube}. For VAE and latent diffusion training, we use 4 NVIDIA A40 GPUs.
The training objectives for the two VAEs—dense and sparse geometry—are unified as:
{\begin{equation}
\small
\begin{array}{ll}
\mathcal{L}_{\mathrm{VAE}} & =\mathbb{E}_{\{\mathcal{G},A_N\}}[\mathbb{E}_{X\sim p({\mathcal{E}})}[\lambda_0\mathrm{BCE}(\mathcal{G},\tilde{\mathcal{G}})+ \\
 & \lambda_1\mathcal{L}_{1}(A_N,\tilde{A}_N)]+ 
 \lambda_2\mathbb{K}\mathbb{L}(p_{\mathcal{E}}(X)\parallel p(X))],
\end{array}
\end{equation}}where \(\mathrm{BCE}(\cdot)\) represents the binary cross-entropy for grid occupancy, \(\mathcal{L}_{1}\) denotes the L1 loss, and \(\mathbb{K}\mathbb{L}(\cdot \parallel \cdot)\) is the KL divergence. In practice, we set the weighting factors to be \(\lambda_0 = 20,\lambda_1 = 50,\lambda_2 = 0.03\). We train the VAE for $\mathcal{G}_d$ and $\mathcal{G}_s$ with 28120 and 71660 iterations, respectively.
The training procedure for 3D latent diffusion follows the same structure implemented by~\cite{ren2024xcube}. For raw input conditioning, we use an additional point encoder to quantize the incomplete point cloud to a voxel grid and concatenate it with the latent feature. We set the diffusion step to be 100 with a linear noise schedule, and trained for 6500 iterations.

\noindent\textbf{Autoregressive transformer} is implemented based on MeshAnything V2~\cite{chen2024meshanythingv2} to minimize the negative log-likelihood of the ground-truth mesh token sequence conditioned on the input point cloud. The training loss is defined as the cross-entropy loss over the entire mesh token sequence:
{\setlength\abovedisplayskip{4pt}
\setlength\belowdisplayskip{4pt}
\begin{equation}
\small
\mathcal{L} = - \sum_{t=1}^{L} \log P\left( t_t = \mathcal{T}_M^{(t)} \mid \mathcal{T}_P, \mathcal{T}_M^{<t}; \phi \right),
\end{equation}}where \( L \) is the length of the target mesh token sequence, \( \mathcal{T}_M^{(t)} \) is the ground-truth token at position \( t \), \( \mathcal{T}_M^{<t} \) denotes all tokens before \( t \), \( \mathcal{T}_P \) is the tokenized point cloud condition, and \( \phi \) represents the parameters of the decoder-only transformer.
We implement our model using the OPT-350M architecture as the backbone transformer. The model is trained for 520000 iterations on 4 NVIDIA A40 GPUs with a total batch size of 8. We freeze the point cloud encoder, which has been trained on large-scale daily object datasets and thus is capable of handling complex geometries.

\subsection{Training Data Processing}


\noindent\textbf{Normalization.} To ensure consistent scale and orientation across all training samples, we implement a robust normalization procedure. The point cloud is first centered by translating its centroid to the origin of the coordinate system. The translation vector is computed as the arithmetic mean of all point coordinates. Following translation, we apply uniform scaling based on the maximum extent along any coordinate axis, normalizing the point cloud to fit within the [-1, 1] range in all dimensions. This normalization strategy preserves aspect ratios while eliminating scale variations that could impede model learning.

\noindent\textbf{Normal Estimation.} We compute surface normals using a k-nearest neighbors approach with 30 neighboring points, providing crucial information about local surface orientation. These normals are essential for capturing fine geometric details and enabling more meaningful latent space representations in the VAE.

\noindent\textbf{Multi-Resolution Voxelization.} The core of our preprocessing pipeline involves converting the normalized point clouds into structured volumetric representations. We employ a sparse voxelization approach that efficiently represents 3D space while minimizing memory requirements. The process begins with an initial voxelization at low resolution of $128^3$, which serves as the foundation for generating multiple resolution levels, then moves to higher resolutions of up to $512^3$.
For each target resolution (\ie, $128^3$ and $512^3$), the voxel centers are transformed to world coordinates and normalized to match the convolutional occupancy networks scale convention, ensuring compatibility with standard 3D deep learning architectures. This multi-resolution approach enables training of hierarchical models that can capture both global structure and local details.

\noindent\textbf{Feature Splatting and Aggregation.} To transfer point-level features to the voxel grid, we implement trilinear splatting operations. This technique distributes point features (normals) to neighboring voxels using trilinear interpolation weights, creating smooth feature fields that preserve spatial relationships. The splatted features are subsequently normalized to unit length, ensuring numerical stability during network training. This process results in two key data components: voxelized geometry and surface normal fields.

\section{Metric Calculation Details}

Our evaluation encompasses both the final mesh and the intermediate point cloud outputs. It is noteworthy that the accuracy metrics for the mesh (such as Chamfer Distance) are derived from point clouds sampled from its surface. Furthermore, the assessment of both point clouds and meshes involves a series of heuristic coordinate alignment transformations to ensure the fairness of the computations.

\subsection{Data Processing}
\textbf{Mesh Preprocessing and Surface Sampling.}
The process begins with point cloud sampling from the mesh surface. To generate a uniformly distributed point cloud from the three-dimensional mesh, we employ a sampling strategy based on face area weighting.
Initially, each face is examined; those with fewer than three vertices or invalid indices are disregarded. For polygonal faces, a fan triangulation method is applied to subdivide them into multiple triangles, ensuring subsequent sampling is performed on standard triangular elements.
For each valid triangle, its area is computed and used as the weighting factor for sampling. The cumulative area of all triangles constitutes the total surface area, which in turn determines the probability of each triangle being selected. This area-weighted approach ensures that larger triangles have a higher likelihood of being sampled, thereby promoting an overall uniform distribution of points across the mesh surface and preventing local density variations due to over- or under-representation of small faces.

After establishing the triangle selection probabilities, each sampling iteration randomly selects a triangle, and a point is generated within its interior using uniformly distributed barycentric coordinates. For each sampled point, two random numbers are generated and subjected to a necessary folding operation to guarantee that the point lies within the triangle boundaries rather than outside. This process is repeated until the desired number of points is attained, resulting in a uniformly distributed point cloud on the mesh surface.

In cases where the mesh contains no valid faces or has an insufficient total surface area, the method defaults to randomly selecting points directly from the vertex set. This fallback mechanism ensures the robustness of the sampling algorithm and prevents failures when processing degenerate or empty meshes. Through this area-weighted sampling strategy, high-quality, uniformly distributed, and highly representative point cloud data can be generated from complex three-dimensional meshes.


\noindent\textbf{Alignment Procedure.} To obtain a robust initial pose, we first center the source and target point clouds by translating them to their respective centroid coordinate systems. Subsequently, we compute the covariance matrices for both centered point clouds and perform eigen-decomposition to extract the Principal Component Analysis (PCA) principal directions as an intrinsic coordinate system representing the shape. After arranging the eigenvectors in descending order of their corresponding eigenvalues and ensuring right-handed coordinate system consistency, we construct a rotation matrix that aligns the PCA coordinate system of the source point cloud to that of the target point cloud, using the target's eigenvectors as reference. The rotation matrix is further regularized via Singular Value Decomposition (SVD) to satisfy orthogonality constraints and maintain a positive unit determinant. The translation vector is then derived by matching the centroid positions of the two point clouds, thereby forming the initial rigid transformation matrix. This PCA-based initialization effectively reduces the search space caused by rotational and scale inconsistencies, enabling subsequent iterations to converge from a more reasonable starting state.

Following initial alignment, we refine the transformation using the Iterative Closest Point (ICP) framework implemented in Open3D, which is scale-sensitive. To enhance robustness, we first estimate normal vectors for both point clouds and employ adaptive thresholds that dynamically adjust the nearest-neighbor search range based on the scale of the point clouds. During iterative optimization, we prioritize the point-to-plane error metric to mitigate noise interference and revert to point-to-point ICP when convergence is insufficient, thereby ensuring algorithmic stability. Finally, by applying the calculated transformation to the predicted point cloud, we obtain a result aligned with the target point cloud.

\subsection{Formulations}
\begin{itemize}
    \item \textbf{Number of vertices, faces, and planes (\#V, \#F and \#P)}
    The number of vertices and faces is determined through direct counting upon loading the meshes. However, extracting geometrically distinct planes requires an effective algorithm.
    To identify distinct planes within the mesh, we begin by extracting the plane parameters corresponding to each face. For a given face, three vertices are selected, and two edge vectors are constructed, from which the face normal vector is derived. If the magnitude of this normal vector is below a negligible threshold, the face is considered degenerate and skipped. For valid faces, the normal vector is normalized to obtain a consistent and stable unit normal. Combined with an arbitrary vertex from the face, the plane offset is computed, thereby forming a complete geometric description of the plane associated with the face.
    After generating the plane parameters for all faces, we proceed to determine whether these planes geometrically represent the same planar surface. Specifically, each newly generated plane is iteratively compared against all previously recorded planes. The comparison first examines whether the normal vectors of the two planes point in the same or opposite direction—if the normals are highly aligned, the planes are considered parallel. Only under the condition of parallelism is the offset difference between the two planes evaluated. If this difference falls within a predefined threshold, the two faces are regarded as lying on the same geometric plane, and the new plane is not counted as distinct. If no matching plane is found within the existing set, the current plane is added as a new unique plane.

    \item \textbf{Failure Rate (FR \%)}
    
    \textbf{City3D}: The determination of a successful mesh is based on a twofold criterion, as implemented in the final validation step of its official repository. The process is deemed successful only if the core reconstruction algorithm returns a positive status and the resulting 3D model contains at least one geometric facet. Specifically, a boolean status flag is first obtained from the reconstruct method, which signifies whether the internal optimization process, incorporating the novel roof preference energy term and hard constraints for topology, converged to a valid solution. Subsequently, the resulting Map data structure is checked for geometric content; a non-zero number of facets confirms that a tangible polygonal mesh was generated. If both conditions are satisfied, the model is serialized to disk. Conversely, if either the algorithm reports a failure (status is false) or produces an empty model, the entire reconstruction is considered to have failed, triggering a runtime exception. 
    
    \textbf{Point2Building} employs an iterative, validation-driven generation process to ensure the structural plausibility of reconstructed meshes (refers to Appendix B in~\cite{liu2024point2building}). Upon generating a candidate set of vertices, the model proceeds to generate the connecting faces. The resulting mesh hypothesis is then subjected to a series of hard-coded geometric checks. These include verifying the presence of a stop token, assessing the coverage of the input point cloud's footprint by the ground polygon, ensuring proper connectivity between walls and the ground edges, and checking for invalid diagonal edges on walls. A mesh is accepted only if it passes all checks. 

    \textbf{BuildAnyPoint}: 
    The post-generation validation process starts by verifying completion through a stop token indicator and identifying any invalid faces containing NaN values. The validated mesh is then saved, labeled with a success or failure designation based on the initial completion check.

    \item \textbf{Chamfer Distance (CD)} between predicted ($A$) and ground truth ($B$) point clouds using the L1 (Manhattan) distance metric. Both point clouds are first uniformly sampled to a fixed size of 16384 points to ensure consistent comparison:
    \begin{equation}
    \small
    \begin{array}{ll}
    CD(A,B)& = \frac{1}{|A|} \sum_{a \in A} \min_{b \in B} \|a - b\|_1 + \\
    &
    \frac{1}{|B|} \sum_{b \in B} \min_{a \in A} \|b - a\|_1 
    \end{array}
    \end{equation}

    \item \textbf{F-score} is computed to evaluate the consistency between two point clouds. Both point clouds are uniformly sampled to a fixed number similar to the calculation of CD. For a predefined distance threshold $d$, precision $P(d)$ is defined as the proportion of points in $A$ whose nearest neighbor in $B$ lies within $d$, while recall $R(d)$ is the proportion of points in $B$ whose nearest neighbor in lies within $d$. The F-score is the harmonic mean of precision and recall:

\begin{align}
\small
\begin{array}{lll}
P(d) &= \frac{1}{|A|} \sum_{a \in A} \mathbb{1}\left( \min_{b \in B} \|a - b\| < d \right) \\
R(d) &= \frac{1}{|B|} \sum_{b \in B} \mathbb{1}\left( \min_{a \in A} \|b - a\| < d \right) \\
F\text{-}score(d) &= \frac{2 \cdot P(d) \cdot R(d)}{P(d) + R(d)}
\end{array}
\end{align}where $\mathbb{1}(\cdot)$ is the indicator function, which returns 1 if the condition is true, and 0 otherwise.

    \item \textbf{Uniformity} quantifies how evenly points are distributed across a surface by measuring the local density variation. A lower score indicates a more uniform distribution, which reflects better surface coverage:
    \begin{equation}
    \small
U(A, r) = \frac{1}{K} \sum_{i=1}^K \left( \frac{n_i - \bar{n}}{\bar{n}} \right)^2
\end{equation} where $r$ denotes the ball radius (set to 0.05) for local density estimation and $K$ represents random query points (set to 100) that are sampled from point cloud $A$.

    \item \textbf{Earth Mover's Distance (EMD)} measures the minimum amount of “work” required to transform one point cloud into another by moving points, where work is defined as the total distance points are moved. It provides a more geometrically faithful comparison than CD by considering the underlying distribution and preserving structural continuity:
    \begin{equation}
    \small
\text{EMD}(A, B) = \min_{\phi: A \to B} \frac{1}{|A|} \sum_{a \in A} \| a - \phi(a) \|_2
\end{equation} where $\phi: A \to B$ is a bijection mapping each point in \( A \) to a unique point in \( B \). 
    
\end{itemize}

\section{More About Results}
\cref{tab:mesh_main} presents the averaged results of our fine-tuned models. Specifically, we performed an additional test/train split on the test set datasets of three scenarios for fine-tuning due to limited computational budget. Our BuildAnyPoint was fine-tuned on this sub-training set, using pairs consisting of diffusion model outputs and ground-truth meshes. Point2Building and City3D were also evaluated under the same subdivided setting. To fully demonstrate the effectiveness of fine-tuning, we additionally report experimental results without fine-tuning in~\cref{tab:mesh_main_suppl}, \ie, where the models were not adapted to intermediate point cloud output and only trained on GT point-mesh pairs. 
The results in \cref{tab:mesh_main_suppl} are obtained using the full test set (around 10,000 for each scenario), in contrast to the subdivided set in \cref{tab:mesh_main} (around 3,000 instances in total). This distinction is driven by the fact that fine-tuning on the entire dataset would necessitate training on an additional 120,000 instances (around 40,000 for each scenario), 
a process that is computationally prohibitive in terms of time. The full fine-tuning results will therefore be provided upon the release of our model weights.
%

\begin{table*}[t!]
\caption{\textbf{Quantitative comparison on structured mesh abstraction for all scenarios without fine-tuning.} Although the absence of a fine-tuning procedure leads to slightly over-tessellated outputs, our method still achieves the best performance in precision-related metrics.}
\vspace{-2mm}
\label{tab:mesh_main_suppl}
\centering
\small
\setlength{\tabcolsep}{1.5mm}{
\scalebox{1}{
\begin{tabular}{l|ccccc|ccccc|ccccc}
\toprule
 &  \multicolumn{5}{c}{LiDAR Point Cloud} & \multicolumn{5}{c}{SfM Point Cloud} & \multicolumn{5}{c}{Sparse Sample Point Cloud}\\
\cmidrule(lr){2-6}
\cmidrule(lr){7-11}
\cmidrule(lr){12-16}
 Methods & \# V $\downarrow$ & \# F $\downarrow$ & \# P $\downarrow$ & FR $\downarrow$ &CD $\downarrow$ & \# V $\downarrow$ & \# F $\downarrow$ & \# P $\downarrow$ & FR $\downarrow$ &CD $\downarrow$ & \# V $\downarrow$ & \# F $\downarrow$ & \# P $\downarrow$ & FR $\downarrow$ &CD $\downarrow$ \\
\midrule
City3D~\cite{huang2022city3d}   & 113 & 42 & \textbf{14} & 1 \% & 0.064 & 221 & 93 & \textbf{16} & 2 \% & 0.118 & 180 & 78 & \textbf{13} & 16 \% & 0.342\\
Point2Building~\cite{liu2024point2building} & \textbf{20} & \textbf{34} & 18 & 1 \% & 0.053 & \textbf{21} & \textbf{36} & 18 & 2 \% & 0.053 & \textbf{19} & \textbf{33} & 16 & 1 \% & 0.044\\
\midrule
Ours (w/o fine-tune) & 42 & 77 & 28 & \textbf{0 \%} & \textbf{0.041} & 37 & 68 & 23 & \textbf{0 \%} & \textbf{0.029} & 37 & 69 & 24 & \textbf{0 \%} & \textbf{0.034} \\
\bottomrule
\end{tabular}}
}
\captionsetup{justification=justified}
\vspace{-3mm}
\end{table*}

\begin{figure*}[th!]
  \centering
   \includegraphics[width=0.96\linewidth]{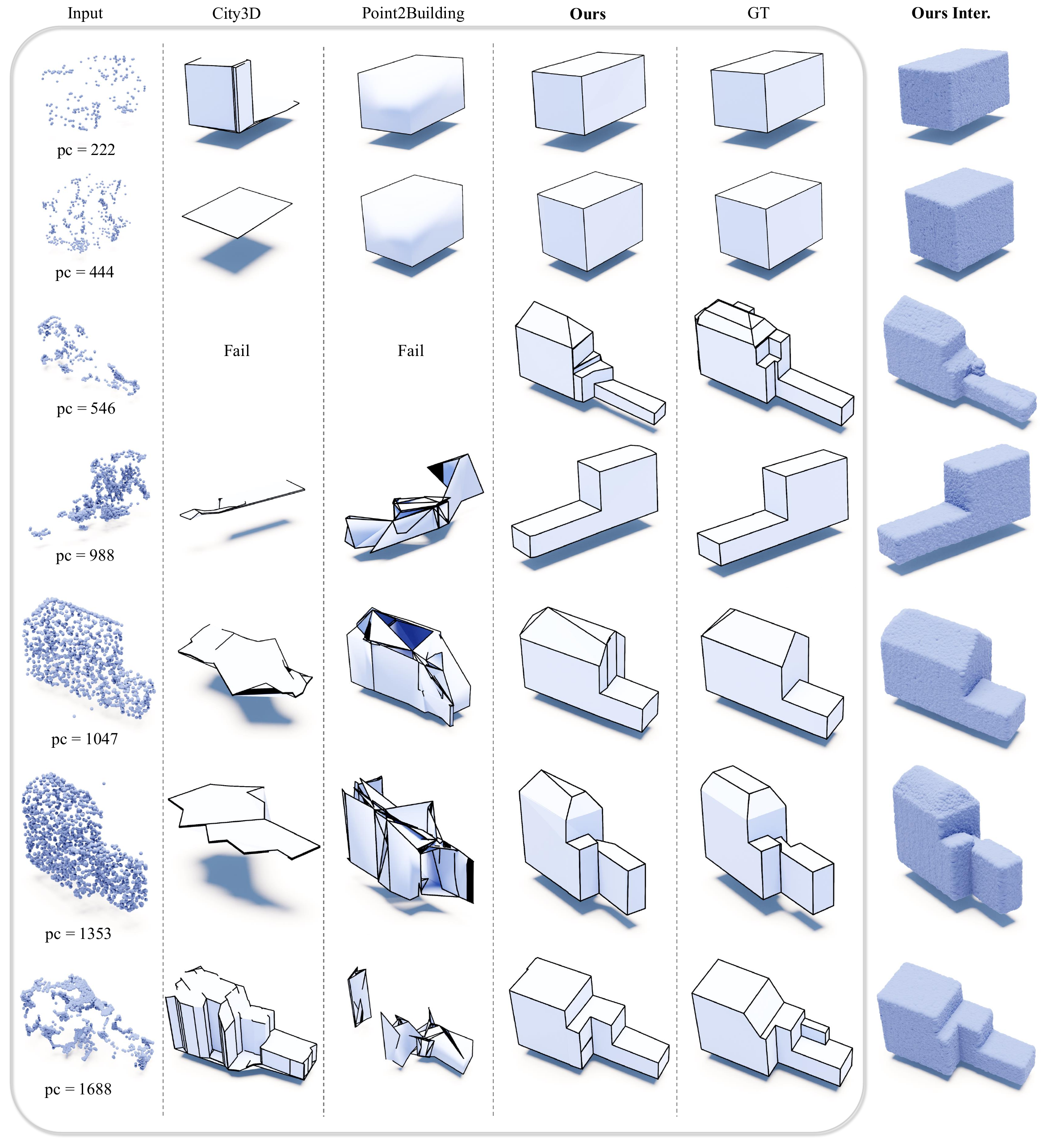}
   \vspace{-4mm}
   \caption{\textbf{More Results.} Our method consistently produces high-quality outputs, regardless of the input density.}
   \label{fig:more-vis}
   \vspace{-4mm}
\end{figure*}


\end{document}